\documentclass[a4paper,fleqn]{cas-dc}

\usepackage{fancyhdr}
\usepackage{lastpage}
\usepackage[numbers,sort&compress]{natbib}

\usepackage{verbatim}     
\usepackage{booktabs}
\usepackage{microtype}

\usepackage{hyperref}     
\usepackage{url}

\usepackage{graphicx} 
\graphicspath{{images/}} 

\usepackage[caption=false,font=footnotesize]{subfig}

\usepackage{amsmath}
\usepackage{mathtools}
\usepackage{amssymb}
\usepackage{bm}
\usepackage{dsfont}
\usepackage{cases}
\mathtoolsset{showonlyrefs}

\usepackage[ruled,noend]{algorithm2e}
\SetKw{Break}{break}

\usepackage{enumitem}

\def\tsc#1{\csdef{#1}{\textsc{\lowercase{#1}}\xspace}}
\tsc{WGM}
\tsc{QE}
\tsc{EP}
\tsc{PMS}
\tsc{BEC}
\tsc{DE}

\begin{document}
\let\WriteBookmarks\relax
\def\floatpagepagefraction{1}
\def\textpagefraction{.001}
\shorttitle{Centralized Collision-free Polynomial Trajectories and Goal Assignment for Aerial Swarms}
\shortauthors{Benjamin Gravell and Tyler Summers}

\title[mode=title]{Centralized Collision-free Polynomial Trajectories and Goal Assignment for Aerial Swarms}                      
\tnotemark[1]

\tnotetext[1]{This material is based on work supported by the United States Air Force Office of Scientific Research under award number FA2386-19-1-4073. This material was also supported by the Army Research Office and was accomplished under Grant Number: W911NF-17-1-0058.}

\author[1]{Benjamin Gravell}[auid=000, bioid=1, orcid=0000-0003-3302-0166]
\cormark[1]
\ead{benjamin.gravell@utdallas.edu}

\address[1]{Department of Mechanical Engineering, The University of Texas at Dallas, Richardson, TX, 75080 USA}

\author[1]{Tyler Summers}[auid=001, bioid=2, orcid=0000-0002-0113-8912]
\ead{tyler.summers@utdallas.edu}

\cortext[cor1]{Corresponding author}

\nonumnote{ Conflict of interest - none declared. }

\begin{abstract}
Computationally tractable methods are developed for centralized goal assignment and planning of collision-free polynomial-in-time trajectories for systems of multiple aerial robots. The method first assigns robots to goals to minimize total time-in-motion based on initial trajectories. By coupling the assignment and trajectory generation, the initial motion plans tend to require only limited collision resolution. The plans are then refined by checking for potential collisions and resolving them using either start time delays or altitude assignment. Numerical experiments using both methods show significant reductions in the total time required for agents to arrive at goals with only modest additional computational effort in comparison to state-of-the-art prior work, enabling planning for thousands of agents.
\end{abstract}


\maketitle

\thispagestyle{empty}

\section{Introduction}

In the burgeoning field of autonomous robotics, aerial robots are quickly becoming a useful platform for firefighting, police, search-and-rescue, surveillance, and product delivery. In the deployment of large fleets of such robots, trajectory plans must satisfy the competing criteria of safety and performance. Specifically, the safety requirement of avoiding vehicle-to-vehicle collisions and the performance requirement of minimizing time-in-flight are considered. As the number of robots increases, so too does the complexity of satisfying these requirements, warranting the development of computationally tractable methods to this end.

A wide class of traditional single-agent motion planning methods rely on discretization of the state space and definition of a related state-transition graph \cite{svestka1998}. Optimal feasible paths are then found through a graph search \cite{dijkstra1959, hart1968, wang2012, koenig2002, koenig2004, stentz1993} or other combinatorial solvers \cite{lavalle2006}. While these methods can solve the multi-agent planning problem e.g. as in \cite{turpin2014ar,sharon2015conflict, adler2015, solovey2015, honig2018}, they become computationally intractable quickly as the number of agents increases, leading to an exponential growth of the search space dimensionality \cite{erdmann1986, lavalle2006}. Some methods have been explored which reduce the search space dimensionality \cite{lavalle2006, wagner2012, wagner2015, solovey2015finding}, but are unable to sufficiently reduce the complexity for large numbers of agents \cite{turpin2014}. Other centralized planning approaches based on sequential mixed-integer linear optimization \cite{schouwenaars2001mixed, richards2002aircraft, ayuso2016}, sequential convex programming \cite{gglr2012, chen2015}, semidefinite programming \cite{frazzoli2001}, formation space-based velocity planning \cite{kloder2006}, or path improvement using revolving areas \cite{solomon2018motion} can work well for relatively small teams but do not scale well to large teams due to high computational complexity.

A complementary approach for motion planning is to use local decentralized feedback control laws to satisfy collision avoidance and dynamic constraints \cite{warren1990multiple,fiorini1998,berg2008,berg2011,guy2009,guy2010,cap2014,cap2015,panagou2020}. These require real-time sense-and-avoid capabilities, increasing the system cost and complexity. In general, these schemes lead to difficulties maintaining optimal or near-optimal performance and avoiding undesirable deadlock situations.

It was recently shown by \cite{turpin2014} that when the robots are interchangeable, combining the assignment and planning problems facilitates finding collision-free trajectories. Therein a concurrent assignment and trajectory planning algorithm was proposed which tractably gave collision-free trajectories for large robot teams for sufficiently spaced start and goal locations. { The trajectories generated by that approach are globally optimal with respect to a total \emph{squared distance} metric and under the assumption of synchronized robot motion}, which resulted in trajectories that can be significantly suboptimal in terms of total time-in-motion and may violate minimum velocity constraints associated with certain aerial robots.
Additionally, it was asserted that coupling the assignment and trajectory planning was critical to avoid overwhelming computational complexity.

To overcome these shortcomings, in this work a similar problem setup and centralized planning setting to \cite{turpin2014} is considered, but allows non-simultaneous trajectory end times and attempts to minimizes the total \emph{time-in-motion} via a strategy with partial coupling between goal assignment and trajectory planning. The proposed algorithm designs piecewise polynomial-in-time trajectories with physical feasibility and computational tractability guarantees; physical systems track the generated trajectories without colliding with each other, while the polynomial formulation enables key calculations such as the point-of-minimum-clearance to be executed via well-studied and efficient root finding algorithms.
It is demonstrated that the proposed method significantly reduces total time-in-motion relative to \cite{turpin2014} with only modest additional computational expense which becomes negligible for large numbers of agents. 
The authors' prior work \cite{Gravell2018} presented a similar approach, but required a simple kinematic model which assumed an unrealistic ability to instantaneously change positions vertically and instantaneously change velocities horizontally; these assumptions are removed in the present work. Further, more extensive simulations and an experimental implementation on a multi-robot testbed are provided.

The remainder of the paper is organized as follows.
First a variation of the trajectory planning problems given in \cite{turpin2014,Gravell2018} is defined, then a strategy in which goal assignment is performed based on initial trajectory plans is proposed, followed by a refinement step to resolve potential collisions using either time delays or altitude assignment. 
Throughout, constraints on the magnitude of time derivatives of position (speed, acceleration, jerk, etc.) are honored.
Finally, simulated and physical experiments on a many-member quadrotor platform are presented that illustrate the effectiveness of the algorithms.

\section{Preliminaries}
{ This work begins by establishing mathematical and notational preliminaries; where applicable, the notation of \cite{turpin2014} is followed.}
Unlike the authors' previous work \cite{Gravell2018} which restricted agents to 2-dimensional planes and assumed instantaneous switching between these planes, now a fully 3-dimensional Euclidean space is considered and accompanying trajectory generation algorithms developed which support implementation on a physical robotic system. { To be clear, in this work all objects and geometry with physical extent reside within a common 3-dimensional Euclidean space.}
The set of integers between $1$ and positive integer $Z$ is denoted by $\mathcal{I}_Z\equiv\{1,2,\dots,Z\}$, and the $Z \times Z$ identity matrix is denoted by $I_Z$. The following symbols are used for certain operators and objects:
\begin{align*}
    \land: &\text{ Logical ``and"} &\lor: &\text{ Logical ``or"}\\
    \cap: &\text{ Set intersection} &\cup: &\text{ Set union}\\
    \emptyset : &\text{ Empty set} &\oplus: &\text{ Minkowski sum}\\
    \text{conv} : &\text{ Convex hull}
\end{align*}
Vectors in $\mathbb{R}^N$ are column vectors unless otherwise specified. The Euclidean norm of such a vector $z$ is denoted as $\| z \|$ and defined as $\| z \| = \sqrt{\sum_{i=0}^N z_i^2}$.
Consider the scenario where $n$ agents begin at $n$ start locations and move towards $n$ goal locations in a 3-dimensional Euclidean space { with a fixed origin and coordinates measured in a Cartesian coordinate system.} The first two coordinates of this space are referred to as ``horizontal'' and the third coordinate as ``vertical''. 

The $i^{th}$ agent position is given by ${x}_i \in \mathbb{R}^3, i \in \mathcal{I}_n$ with the horizontal and vertical parts denoted as ${x}_{i,12} \in \mathbb{R}^2$ and ${x}_{i,3} \in \mathbb{R}^1$ respectively.
The first four derivatives of position with respect to time are called velocity, acceleration, jerk, and snap. These and higher-order derivatives are collectively referred to as time derivatives. { Derivatives with respect to time are notated either by dots or by numbers enclosed by parentheses above the variable, e.g. ${\dot{x}}_i = \overset{\scriptscriptstyle(1)}{x}_{i}$ is velocity and ${\ddot{x}}_i = \overset{\scriptscriptstyle(2)}{x}_{i}$ is acceleration of agent $i$.}

The collision volume of agent $i$ is represented by a finite cylinder $\mathcal{C}_{i}$ of radius $R_i$ and height $H_i$. Each cylinder is centered at ${x}_i \in \mathbb{R}^3$. The cylinders represent the safe collision volume around an agent; a collision occurs if and only if two cylinders intersect. The cylinders have orientations which remain fixed with the axial direction parallel to the world vertical direction (a ``vertical'' cylinder). The variable height of the cylinders relative to the radius is useful for modeling phenomena such as downwash from the rotors of a quadrotor vehicle. Let the largest radius and height of all vehicles be $R = \max_i(R_i)$ and $H = \max_i(H_i)$.

The $i^{th}$ start location and $j^{th}$ goal location are given by ${s}_i \in \mathbb{R}^3, i \in \mathcal{I}_n$ and ${g}_j \in \mathbb{R}^3, j \in \mathcal{I}_n$ respectively. 
The horizontal ground plane is the 2-dimensional set $\{x | x_3=0\}$.
The agents operate in a region $\mathcal{K}$:
\begin{align}
    \mathcal{K} \equiv \text{conv}\left(\{{s}_i|i\in \mathcal{I}_n\} \cup \{{g}_j|j\in \mathcal{I}_n\}\right) \oplus \mathcal{C}_{\infty}. \label{eq:region_K}
\end{align}
where $\mathcal{C}_\infty$ is a vertical cylinder with radius $R$ whose horizontal coordinates of the center are on the origin and with vertical extent from the ground to positive infinity.
An altitude $\mathcal{A}(\eta)$ at height $\eta$ is the subset of $\mathcal{K}$ within which any agent at that height is contained i.e. a 3-dimensional horizontal slab volume:
\begin{align}
    \mathcal{A}(\eta) &= \mathcal{K} \cap  \{x : | x_3-\eta | \leq H/2 \}
\end{align}
Define the $n\times 3$ goal matrix as 
\begin{align}
    {G}=
    \begin{bmatrix}
    {{g}_1} & {{g}_2} & \dots & {{g}_n}
    \end{bmatrix}^\intercal.
\end{align}
Define the $n\times n$ boolean assignment matrix $\phi$, which assigns agents to goals, as
\begin{equation} 
\phi_{ij} = 
\begin{cases}
1 & \text{if agent } i \text{ is assigned to goal } j\\
0 & \text{otherwise}
\end{cases}
\end{equation}
Therefore row $i$ of $\phi{G}$, denoted as $(\phi{G})_i$, gives the goal location assigned to agent $i$.
All agents are assigned to goals in a one-to-one mapping so
\begin{equation} \label{eq:assignment_constraint}
\phi^\intercal\phi = I_n.
\end{equation}

{ A polynomial $p(t):\mathbb{R}\rightarrow{\mathbb{R}}$ of scalar $t$ with degree $d$, with $d+1$ coefficients $\alpha_i$, is given by
\begin{equation}
p(t) = \alpha_0 + \alpha_1 t + \alpha_2 t^2 + ... + \alpha_d t^d = \sum_{i=0}^d  \alpha_i t^i . \label{eq:polynomial}
\end{equation}
}
Minimizing a polynomial over a finite domain interval $[t_0,t_f]$ is a straightforward and computationally efficient procedure which follows from Fermat's theorem for stationary points from differential calculus. A description is provided in Algo. \ref{algo:polynomial_minimization} for completeness since it will be called at various points throughout this work.
Maximization is accomplished by the same algorithm by passing a negated polynomial. Also, both minima and maxima can be found concurrently at little additional computational expense by a simple modification to the algorithm.

Similarly, finding the intervals of a finite domain interval $[t_0,t_f]$ over which a polynomial evaluates { within a range of values i.e. finding the set}
\begin{align}
    \{t \ | \ p(t) \in [p_-,p_+] \text{ and } t \in [t_0,t_f] \}
\end{align}
is also straightforward and efficient. The procedure begins by finding domain intervals where the polynomial evaluates above the lower limit $p_-$ and below the upper limit $p_+$ (Alg. \ref{algo:polynomial_above}), then intersects those domain intervals with eachother and the prescribed interval $[t_0,t_f]$ (Alg. \ref{algo:polynomial_interval}). Note that Alg. \ref{algo:polynomial_above} works identically for finding intervals where a polynomial is below an upper value $p_+$ by simply reversing inequalities. 

{ The trajectory planning problem and the proposed solution techniques are now introduced.}

\section{Trajectory Planning Problem}
{ The trajectory planning problem requires finding $n$ instances of $3$-dimensional trajectories which guide $n$ agents from start to goal locations. The trajectories are given agent-wise by}
\begin{equation*}
\gamma_i(t): [t_{0,i}, t_{f,i}] \rightarrow {x}_i,\quad i \in \mathcal{I}_n
\end{equation*}
and must satisfy the initial and terminal conditions
\begin{align}
\gamma_i(t_{0,i})&={s}_i,\quad i \in \mathcal{I}_n \label{eq:init_position_constraint}, \\
\gamma_i(t_{f,i})&=\left(\phi{G}\right)_i,\quad i \in \mathcal{I}_n \label{eq:term_position_constraint}.
\end{align}
Agents are considered to be quadrotors, whose center position dynamics linearized about the hover configuration are modeled as a quadruple integrator in horizontal directions (due to the rolling action which must precede lateral acceleration) and a double integrator in the vertical direction \cite{mellinger2011}:
\begin{align} \label{eq:dynamics_constraint}
    \overset{\scriptscriptstyle(4)}{x}_{i,12}(t) = u_{\text{horz,i}} \ , \quad \overset{\scriptscriptstyle(2)}{x}_{i,3}(t) = u_{\text{vert,i}}
\end{align}
where $u_{\text{horz,i}}$ and $u_{\text{vert,i}}$ are control inputs. The dynamics are not used explicitly in terms of designing control inputs, but rather are used to motivate the choice of trajectory form, namely piecewise polynomials of a particular order.

{ By choosing a whole number $q$ sufficiently high and imposing constraints on the norm of $q-1$ time derivatives, actuator constraints are honored. The particular choice for $q$ in the case of quadrotors is established in Section \ref{subsec:base_polynomial}.}
These constraints are encoded in a vector $\delta \in \mathbb{R}^{q-1}$ with $\delta_k > 0$ and applied as
\begin{align} \label{eq:derivative_constraint}
    \| \overset{\scriptscriptstyle(k)}{\gamma}_i(t) \| \leq \delta_k \text{ for } k \in \mathcal{I}_{q-1}.
\end{align}
Define the global start and end times for which motion may occur over all agents:
\begin{equation*}
\begin{aligned}
t_{0,\text{all}}&=\text{min}(t_{0,1},t_{0,2},\dots,t_{0,n}), \\
t_{f,\text{all}}&=\text{max}(t_{f,1},t_{f,2},\dots,t_{f,n}).
\end{aligned}
\end{equation*}
Ensure collision avoidance by requiring the collision volumes of all agent pairs to be disjoint during the period of possible motion:
\begin{align} \label{eq:collision_free_constraint}
\{x_i(t) \oplus \mathcal{C}_i\} \cap \{x_i(t) \oplus \mathcal{C}_j\} = \nonumber \emptyset \\
\text{ for } t:~[t_{0,all},t_{f,all}],\quad i\neq j \in \mathcal{I}_n. 
\end{align}
Like the previous work in \cite{Gravell2018}, the proposed method aims to minimize the total, or equivalently average, time-in-flight of all agents. This is a useful cost metric for many applications e.g. product delivery and emergency response. Therefore, the optimization problem seeks trajectories $\gamma^*(t) = [\gamma_1(t),...,\gamma_n(t)]$ and goal assignment $\phi^*$ that minimize total time-in-motion:
\begin{equation} \label{eq:opt_prob_orig}
\begin{aligned}
& \gamma^*(t), \phi^* = &&\underset{\gamma(t), \phi}{\text{argmin}} 
\sum_{i=1}^n \int_{t_{0,i}}^{t_{f,i}} dt \\
&&& \text{subject to }
\eqref{eq:assignment_constraint}, \eqref{eq:init_position_constraint}, \eqref{eq:term_position_constraint}, \eqref{eq:dynamics_constraint},\eqref{eq:derivative_constraint},\eqref{eq:collision_free_constraint}
\end{aligned}
\end{equation}
{
\textbf{Assumptions:} The following assumptions are explicitly imposed as part of the problem formulation:
\begin{enumerate}[label=(A\arabic*)]
    \item Any assignment of agents to goals is permissible.
    \item The collision volume of each agent is the set of points contained in cylinder $\mathcal{C}_{i}$.
    \item The effect of any dynamics model mis-specification, imperfect state knowledge, actuation error, and external disturbance are small enough such that the true physical extent of each agent is always fully contained inside the collision volume $\mathcal{C}_{i}$.    
    \item Continuity and satisfaction of upper bound constraints on $q-1$ time derivatives of position is sufficient to ensure actuator constraints are honored.
    \item The region $\mathcal{K}$ in \eqref{eq:region_K} is devoid of any obstacles other than the agents themselves.
    \item The region $\mathcal{K}$ in \eqref{eq:region_K} has infinite positive vertical extent.
    \item All start and goal locations are fixed on a common ground plane and are spaced at least $2R$ apart:
    \begin{align*}
        s_{i, 3} = g_{i, 3} &= 0 \ \forall \ i \in \mathcal{I}_n \\
        \| s_i - s_j \| &> 2R \ \forall \ i \neq j \in \mathcal{I}_n \\
        \| g_i - g_j \| &> 2R \ \forall \ i \neq j \in \mathcal{I}_n
    \end{align*}
\end{enumerate}
}
{
The modeling assumption of no uncontrolled obstacles in the operating space is not altogether unreasonable when considering the nearly empty airspace encountered at altitudes above tree tops, buildings, etc. in typical real-world flight scenarios. The use of cylindrical collision volumes renders orientation of each quadrotor irrelevant for the purpose of trajectory planning. 
}

{ The solution to the global problem in \eqref{eq:opt_prob_orig} is ultimately not obtained exactly, but rather a suboptimal solution is found using \eqref{eq:opt_prob_orig} to guide generation of trajectories and goal assignment by the approach proposed in the following subsections.
The strategy for finding an approximate solution to this problem proceeds by temporarily ignoring the clearance requirements \eqref{eq:collision_free_constraint} which effectively reduces the domain of trajectories under consideration to the ground plane, choosing a function form for trajectories (piecewise polynomials) to reduce the problem to goal assignment, generating horizontal trajectories, then constructing vertical trajectories using refinement techniques which detect and resolve collisions. 
As a result, these trajectories will be shown to be feasible (e.g. collision-free) and computable after a finite number of operations by construction.

As the trajectory generation procedure based on piecewise polynomial functions is used throughout the goal assignment and collision resolution phases, the trajectory generation scheme is described next.}

\subsection{Trajectory Generation} \label{sec:traj}
The trajectory design is motivated by the observation that minimum-time trajectories along a long straight line with maximum speed constraints will naturally partition into three segments; acceleration, constant (max) speed, and deceleration. A similar idea has previously been suggested for point-to-point robot trajectory planning under the name ``Linear Segments with Parabolic Blends" \cite{spong2008}. This idea is generalized to higher-order acceleration (blend) segments. During the acceleration segments, one or more time derivatives of order 2 and higher will be pushed to a constraint maximum, and during the constant max speed segment the higher order time derivatives will be zero. Although physical models involving friction { (i.e. higher fidelity models than that assumed in \eqref{eq:dynamics_constraint})} theoretically allow only asymptotic approach of the maximum speed under actuation constraints e.g. the exponential approach of the speed of a particle in gravitational free-fall to a terminal speed, in practice it was found that the polynomial trajectories were sufficient for reference tracking.

This work does not attempt to optimize control effort during the acceleration segments since the control effort expended during the constant speed segment dominates e.g. due to air friction and by virtue of the relative duration of this segment over long horizontal paths. If deemed necessary, techniques such as minimum-snap trajectory design via quadratic programming \cite{mellinger2011} could be utilized to further decrease the control effort, possibly at the expense of trajectory duration and computational burden. Any techniques which return polynomial-in-time trajectory segments are fully compatible with the the remainder of the proposed method.

{ This work also restricts trajectories to strictly piecewise vertical and horizontal straight-line paths, which permits simplified trajectory planning and collision resolution by treating trajectories as single-dimensional polynomials of time multiplied by a constant unit heading vector. A pair of tuples $\delta_{\text{horz},k}$ and $\delta_{\text{vert},k}$ are used and \eqref{eq:derivative_constraint} is used with $\delta_k$ set to either $\delta_{\text{horz},k}$ or $\delta_{\text{vert},k}$ depending on whether $\overset{\scriptscriptstyle(k)}{\gamma}_i(t)$ is horizontal or vertical at $t$.}

The acceleration segments are individualized polynomials scaled from a base polynomial. The base polynomial is calculated only once at the beginning of the overall routine. Particular whole trajectories are generated by joining acceleration and constant speed segments. { Generation of the base polynomial and individualized polynomials are described in the subsequent two subsections.}

\subsubsection{Base polynomial} \label{subsec:base_polynomial}
{ Recalling the definition of a polynomial of degree $d$ in \eqref{eq:polynomial} and the whole number $q$ which represents the number of time derivatives on which constraints will be enforced, let $2q=d+1$.} It is evident that a given $2q-$tuple of initial and terminal time derivative conditions ($2q$ total point constraints) uniquely specifies a polynomial of degree $d$ so long as the problem is well-posed i.e. if a certain coefficient matrix $A$ is invertible. To ensure continuity of position and $q-1$ time derivatives at the endpoints, specify $q$ constraints at $t=0$ and $q$ constraints at $t=T$. Due to the assumption on the dynamics in \eqref{eq:dynamics_constraint}, by choosing reference trajectories which are piecewise polynomial with degree at least 4 and 2 respectively, open-loop control with sufficient control effort and the absence of disturbances would give perfect tracking. It is also desirable to make the segment transitions smooth to avoid discontinuous control signals. Choosing degree 9 would allow the specification of 5 endpoint time derivative constraints: position, speed, acceleration, jerk, and snap. However, to reduce the computational storage requirement for the trajectories during implementation on actual hardware and reduce computational effort during centralized trajectory planning, a degree of 7 is used. It was found that the difference between the degree 7 and 9 polynomials was extremely slight and in practice the reference tracking error was dominated by other noise sources. 
For comparison, degree 1 polynomials represent constant speed trajectories; this was effectively the approach taken in the authors' previous work \cite{Gravell2018}. The procedure for calculating the base polynomial is as follows:
\begin{enumerate}
    \item   Form the vector of endpoint conditions 
            \begin{align}
                b &=     [p(0),\dot{p}(0),\ldots,\overset{\scriptscriptstyle(q)}{p}(0), \nonumber \\
                  &\qquad  p(T),\dot{p}(T),\ldots,\overset{\scriptscriptstyle(q)}{p}(T)]^\intercal
            \end{align}
    \item   Form the matrix of coefficients $A \in \mathbb{R}^{2q \times 2q}$ as
            \begin{align}
                A_{ij} &=   \left\{\begin{array}{lr}
                            i! & \text{ if } i  =  j \text{ and } i \leq q\\
                             0 & \text{ if } i\neq j \text{ and } i \leq q\\
                             \frac{(j-1)!}{(j-(i-q))!} T^k & \text{ if } i-q \leq j \text{ and } i > q\\
                             0 & \text{ if } i-q > j \text{ and } i > q
                            \end{array}\right.
            \end{align}
            where $k={(j-1)+(i-q-1)}$. This follows from simple differentiation of polynomials and matching coefficients according to the endpoint constraints.
            As an example, for $d=7$ and $T=1$ one has
            \begin{align}
                A = \begin{bmatrix}
                 1  &   0  &   0  &   0  &   0  &   0 &    0    & 0 \\
                 0  &   1  &   0  &   0  &   0  &   0 &    0    & 0 \\
                 0  &   0  &   2  &   0  &   0  &   0 &    0    & 0 \\
                 0  &   0  &   0  &   6  &   0  &   0 &    0    & 0 \\
                 1  &   1  &   1  &   1  &   1  &   1 &    1    & 1 \\
                 0  &   1  &   2  &   3  &   4  &   5 &    6    & 7 \\ 
                 0  &   0  &   2  &   6  &  12  &  20 &   30    &42 \\
                 0  &   0  &   0  &   6  &  24  &  60 &  120   &210 
                \end{bmatrix}
            \end{align}
    \item   Solve the system of linear equations $A\alpha=b$ to obtain the vector of polynomial coefficients $\alpha$.     
\end{enumerate}
In this framework, other polynomial bases such as the orthogonal polynomials of Chebyshev or Legendre could be used to improve the conditioning of the $A$ matrix \cite{Mellinger2012} { i.e. to encourage the singular values of the $A$ matrix to remain clustered around unity and ensure numerical stability of the solution to $A\alpha=b$}; for ever-higher degree polynomials the conditioning of the matrix in the monomial basis degrades. However, for simplicity, monomials are used since the error was found to be manageable on the problem instances encountered, i.e. for degree $7$ polynomials. 
If position and the first $q-1$ time derivatives are $0$ at $t=0$, the first $q$ coefficients $\alpha_0,\ldots,\alpha_q$ are also zero, which is evident from the partial diagonal structure of $A$. Indeed, it is desirable to create an acceleration polynomial which has $p(0) = 0$, $\dot{p}(0) = 0$, $p(T) > 0$, $\dot{p}(T) > 0$ and some higher-order time derivatives zero at both endpoints i.e.
\begin{align}
    b &= [0,0,\ldots,0, x_f,v_f,0,\ldots,0]^\intercal
\end{align}

Although this procedure will always generate a polynomial which satisfies the endpoint constraints, the behavior between the endpoints is governed by the duration $T$. In particular, there is a unique setting of $T$ which ensures that both the position and velocity monotonically increase from the initial to terminal points, thus ensuring that the endpoints are where the minimum and maximum position and speed occur over the segment. This setting is  
\begin{align}
    T=2 \left| \frac{p(T)-p(0)}{\dot{p}(T)-\dot{p}(0)} \right|=2 \left| \frac{p(T)}{\dot{p}(T)} \right|. \label{eq:T_setting}
\end{align}
With this choice, as an additional benefit, the polynomial degree is reduced by 1 i.e. $\alpha_d$ = 0. 
Although proving these facts for arbitrary degree polynomials is difficult, it is now shown that at least for $d=7$, which is the case of interest in this work, that the given setting of $T$ in \eqref{eq:T_setting} gives the desired behavior. Assuming, without loss of generality, that $p(0)=0$, $\dot{p}(0)=0$, $p(T)=0.5$, $\dot{p}(T)=1$, and by \eqref{eq:T_setting} set $T=1$. Solving for the coefficients of the position polynomial obtain
\begin{align}
    p(t)    &= t^6-3t^5+2.5t^4
\end{align}
and differentiating, the acceleration is 
\begin{align}
    \ddot{p}(t) &= 30t^4-60t^3+30t^2 \\
                &= 30 t^2 (t-1)^2
\end{align}
which is nonnegative for all $t$ and thus on the interval $[0,T]$. Thus the velocity monotonically increases from 0 and so does the position, as desired. Attempting to show this for any other setting of $T$ will fail; a proof of this fact is left to future work, noting that a product-of-squares argument (as here) is insufficient to prove a setting of $T$ gives an acceleration which is somewhere negative.


It is emphasized that the base polynomial only needs to be calculated once at the beginning of the overall routine and can be scaled and translated (in time) as necessary for each particular trajectory.
The base polynomials for vertical and horizontal trajectories are calculated separately to account for differing actuation constraints in each direction. In each case, a unit path length and terminal speed equal to the max speed ($p(0)=0$, $\dot{p}(0)=0$, $p(T)=1$, $\dot{p}(T)=\delta_1, T = 2$) are used. This results in the polynomials $p_\text{base,horz}$ and $p_\text{base,vert}$.

\subsubsection{Individualized polynomials}
Once the base polynomials for an acceleration segment have been found, a piecewise polynomial (sub)trajectory may be generated which connects any two points $x_0$, $x_f$ with a straight line path, subject to the time derivative constraints. Let the distance between $x_0$, $x_f$ be $\ell_\Delta = \|x_f-x_0\|$. The whole piecewise polynomial trajectory for agent $i$ with $n_i$ pieces has the form
\begin{align}
    \gamma_i(t) =  \Big\{ p_{ik}(t) \hat{h}_{ik} \text{ if } t \in [t_{ik},t_{ik+1}], k \in \mathcal{I}_{n_i-1}  \Big\}
\end{align}
where $\hat{h}_{ik} \in \mathbb{R}^3$ is a unit heading vector. In this work, this heading will either be horizontal $\hat{h}_{ik} = [a,b,0]^\intercal$ or vertical $\hat{h}_{ik} = [0,0,1]^\intercal$ where $a,b$ are dummy constants satisfying $a^2 + b^2 = 1$. Also, in this work these trajectories are comprised of 2- or 3-segment subtrajectories and 1-segment stationary wait segments. For notational compactness, let 
\begin{align}
    \gamma_{ik} = \{p_{ik},\hat{h}_{ik},[t_{ik},t_{ik+1}]\}
\end{align}
represent a polynomial trajectory segment which encodes a polynomial, a heading, and a time interval.

Accordingly, the norm of the time derivatives has the simplified form
\begin{align}
    \left\| \overset{\scriptscriptstyle(k)}{\gamma}_i(t) \right\| = \Big\{ \left| \overset{\scriptscriptstyle(k)}{p}_{ik}(t) \right| \text{ if } t \in [t_{ik},t_{ik+1}], k \in \mathcal{I}_{n_i-1}
\end{align}

It will be useful to keep in mind the spatial and temporal scaling formulas for polynomials:
\begin{align}
    c p(t) &= c \sum_{i=0}^d  \alpha_i t^i \label{eq:spatial_scale} \\
    p(ct) &= \sum_{i=0}^d  \alpha_i (ct)^i = \sum_{i=0}^d  \alpha_i c^i t^i \label{eq:time_scale} \\
    \frac{1}{c} p(ct) &= c^{-1} \sum_{i=0}^d  \alpha_i (ct)^i = \sum_{i=0}^d  \alpha_i c^{i-1} t^i \label{eq:combine_scale}
\end{align}
from which it follows that the derivatives satisfy:
\begin{align}
    \frac{d^k}{dt^k} (c p(t)) &= c \frac{d}{dt} p(t) \label{eq:spatial_scale_derv} \\
    \frac{d^k}{dt^k} (p(ct)) &= c^k \frac{d^k}{d\tau^k} (p(\tau))  \label{eq:time_scale_derv} \\
    \frac{d^k}{dt^k} \left( \frac{1}{c} p(ct) \right) &= c^{k-1} \frac{d^k}{d\tau^k} (p(\tau)) \label{eq:combine_scale_derv}
\end{align}
where $\tau = ct$.

First, temporal scaling is applied to the acceleration segment in order to ensure the terminal speed is the agent max speed so that
\begin{align}
    | \dot{p}(t) | \leq \delta_1 \text{ for } t \in [0,T]
\end{align}
with equality ensured exactly at $t=T$. 
The (absolute) maximum time derivative $\max_t(| \dot{p}(t) |)$ of the base polynomial is computed via Algorithm \ref{algo:polynomial_minimization} with the interval $[0,T]$. The scale factor is found as $c = \frac{\max_t(| \dot{p}(t) |)}{\delta_1} $ then temporal scaling is applied as
\begin{align}
    p(t) \leftarrow p \left( c t \right) \quad \text{ and } \quad
    T \leftarrow T/c
\end{align}
which achieves the proper scaling of speed per \eqref{eq:time_scale_derv} and preserves the path length traversed.
Next, scaling is applied to the acceleration segment in order to satisfy constraints on the higher time derivatives which are denoted by $\delta \in \mathbb{R}^{q-1}$ so that
\begin{align}
    | \overset{\scriptscriptstyle(k)}{p}(t) | \leq \delta_k \text{ for } t \in [0,T], k \in \mathcal{I}_{q-1}
\end{align}
with equality ensured in at least one derivative at one time. This minimizes the time taken to traverse the path by taking full advantage of the available time derivatives.
The (absolute) maximum time derivatives $\max_t(| \overset{\scriptscriptstyle(k)}{p}(t) |)$ of the base polynomial are computed via repeated applications of Algorithm \ref{algo:polynomial_minimization} with the interval $[0,T]$. Once the (absolute) maximum time derivatives have been identified, scale factors $\psi_k$ associated with satisfying each time derivative constraint are found by
\begin{align}
    \psi_k = \left( \frac{\max_t(| \overset{\scriptscriptstyle(k)}{p}(t) |)}{\delta_k}\right)^{\frac{1}{k-1}} \text{ for } k \in \mathcal{I}_{q-1} \label{eq:scale_factors} .
\end{align}
The maximum of these scale factors is the only one that is needed to ensure all constraints are satisfied, so take $\psi_*=\max_k(\psi_k)$.
The scaling is then applied by 
\begin{align}
    p(t) \leftarrow \psi_* p(t/\psi_*) \quad \text{ and } \quad
    T \leftarrow T \psi_*
\end{align}
which compresses the trajectory temporally and stretches it spatially in equal proportions such that the terminal speed remains the same, per \eqref{eq:combine_scale_derv}, while honoring all higher order time derivative constraints, per \eqref{eq:time_scale_derv}.

Next, a determination of whether a middle constant speed segment is needed is made. This is accomplished by comparing the path length needed by the acceleration segment to reach max speed and (half) the actual path length between the physical endpoints i.e. if $2p(T) < \ell_{\Delta}$ then a constant speed segment is needed. This segment is trivial to calculate; it is simply a constant maximum speed segment whose duration is simply $T_{\text{cs}} = \frac{\ell_{\Delta}-2p(T)}{\delta_1}$. On the other hand, if $2p(T) \geq \ell_{\Delta}$ then no constant speed segment is needed and the acceleration segments must be scaled again to reduce their path length, in which case the maximum speed will not be attained. 

\begin{figure}[pos=h]
\centering
\includegraphics[width=2.5in]{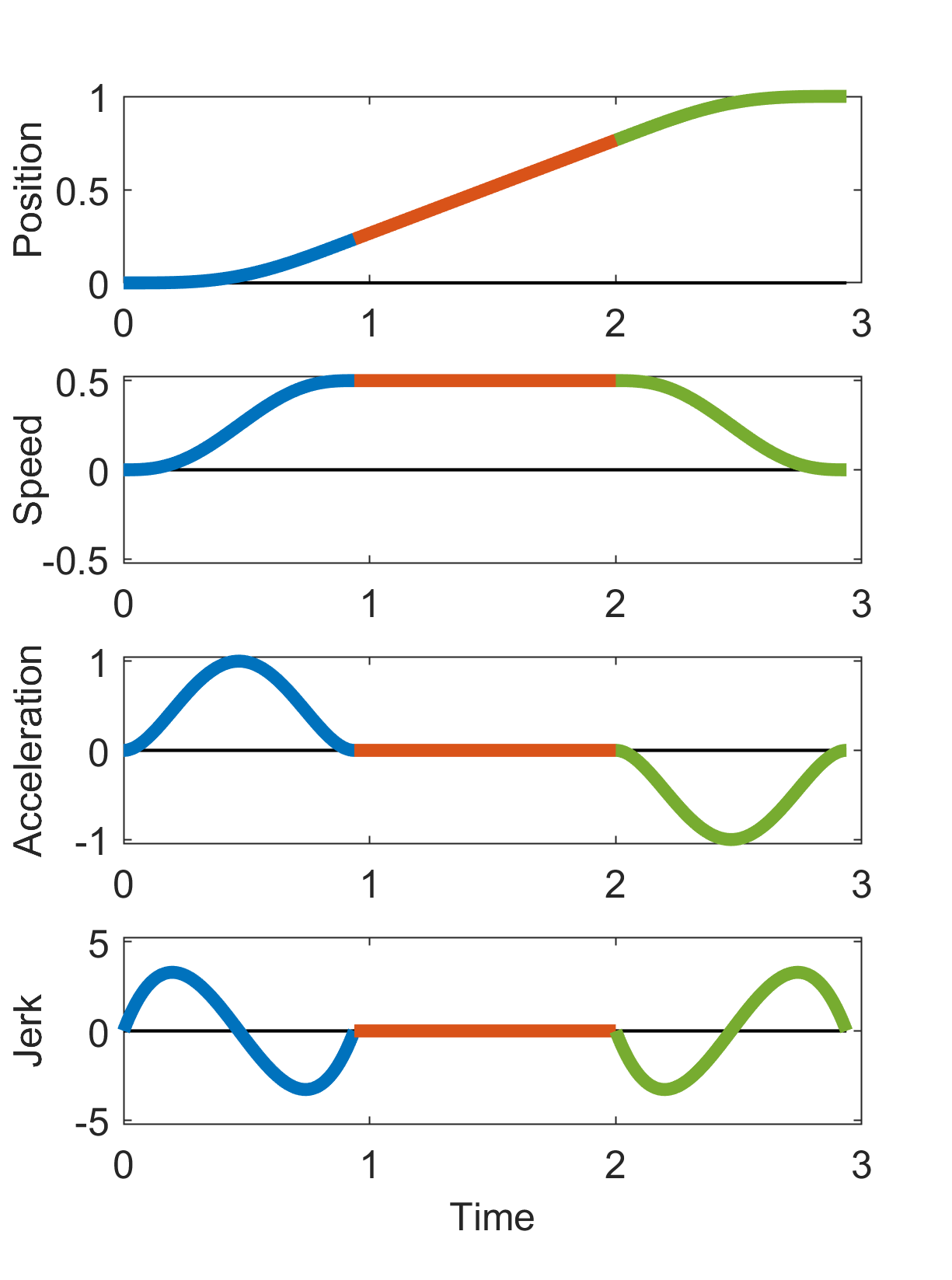}
\caption{Plots of position and its time derivatives for an example piecewise polynomial subtrajectory generated by the proposed method. This example uses degree $d=7$ to accommodate $q-1=3$ time derivative constraints depicted as the upper and lower limits of the vertical axis and tightly bound the speed and at least one higher derivative (in this case acceleration). This example has 3 segments: acceleration, constant speed, and deceleration.}
\label{fig:figure_polytraj1}
\end{figure}

The process continues with a spatial stretch in order to fit the path length exactly:
\begin{align}
    p(t) \leftarrow \frac{\ell_{\Delta}}{2p(T)} p(t)
\end{align}
This has the effect of strictly decreasing the time derivatives per \eqref{eq:spatial_scale_derv} since the scale factor is less than 1.
Then new scale factors are calculated similarly to \eqref{eq:scale_factors} and a temporal stretch is applied to further optimize the trajectory by making full use of the available ``capacity'' of higher order time derivatives.:
\begin{align}
    \psi_k^\prime &= \left( \frac{\max_t(| \overset{\scriptscriptstyle(k)}{p}(t) |)}{\delta_k}\right)^{\frac{1}{k}} \text{ for } k \in \mathcal{I}_{q-1} ,
    \psi_*^\prime =\max_k(\psi^\prime_k) \\
    p(t) &\leftarrow p(t/\psi_*^\prime) , \quad \text{and} \quad
    T \leftarrow T \psi_*^\prime  .
\end{align}
The end result of this entire procedure is a piecewise polynomial (sub)trajectory with 2 or 3 pieces with the first $q$ time derivatives continuous and satisfies all initial, terminal, and range constraints. See Fig. \ref{fig:figure_polytraj1} for an illustrative example.

\subsection{Goal Assignment} \label{sec:goal_assignment}

{
Having described the piecewise polynomial trajectory generation procedure, it is now possible to reduce the problem in \eqref{eq:opt_prob_orig} to one of a linear assignment (combinatorial) goal assignment problem by fixing the functional form of the trajectories.
As in \cite{Gravell2018}, if the collision avoidance constraint is ignored \eqref{eq:collision_free_constraint}, an argument from the calculus of variations shows that trajectories which minimize the integral of $dt$, which is the time-in-motion, follow straight line paths and achieve the highest average speed possible while satisfying the boundary conditions and position derivative constraints.
Thus the problem reduces to simply connecting each start to each goal with minimum-time trajectories on straight line paths, computing the time-in-motion incurred by each trajectory, and finding the goal assignment which minimizes total time-in-motion. If these minimum-time trajectories are replaced with constant velocities, as in \cite{Gravell2018}, the problem amounts to minimizing the total \emph{non-squared} distance. 
Unlike \cite{Gravell2018}, motivated by Section \ref{sec:traj}, we now replace these minimum-time trajectories with piecewise polynomial trajectories, where the expression of the cost in terms of distance is more complicated and is driven by the size of the constraints on the time derivatives (which determine the base polynomial) relative to the distances. 
}

Therefore, the optimal assignment is given by
\begin{equation*}
    \phi^* = \underset{\phi}{\text{argmin}}\sum_{i=1}^{n}\sum_{j=1}^{n} \phi_{ij}C_{ij}
\end{equation*}
where the cost matrix $C$ encodes the cost of assigning agent $i$ to goal $j$. In accordance with \eqref{eq:opt_prob_orig}, $C$ contains the values of the time-in-motion taken by agent $i$ to travel to goal $j$ along a straight line. These times $T_{ij}$ are found by calculating polynomial segment trajectories for agent $i$ moving from start $s_i$ to goal $g_j$ by the procedure described earlier:
\begin{equation*}
C_{ij} = T_{ij},\quad i \in \mathcal{I}_n, j \in \mathcal{I}_n.
\end{equation*}
Due to the exceptionally simple form of the piecewise polynomial trajectories, calculating the $n(n-1)/2$ trajectories for each start-goal pair remains computationally tractable compared with the simplified case of constant velocity trajectories.
This problem may be efficiently solved to optimality with a finite number of iterations using the well-known Hungarian algorithm \cite{kuhn1955,munkres1957}, which runs in $\mathcal{O}(n^3)$ time. Alternate algorithms such as the auction algorithm could also be used with the same time complexity, but with the benefit of parallelization \cite{bertsekas1989,bertsekas1991}.
After solving the optimal assignment, the presumptive horizontal trajectories for each agent are simply chosen as those from the cost matrix generation which are selected by the optimal assignment.

A comparison with the C-CAPT algorithm of \cite{turpin2014}, which uses a cost function of the distance traveled squared, is given in \cite{Gravell2018}. The main disadvantages of the C-CAPT algorithm are that the speed of agents is limited due to the requirement of agents to start and arrive at goals at the same time, as well as a minimum separation spacing between starts and between goals of $2\sqrt{2}R$. The advantage of allowing asynchronous goal arrival is highly dependent on the distribution of the start and goal locations; when some trajectory lengths are much larger than others, the ability to arrive earlier than other agents significantly improves utilization of the available actuation resources, e.g. speed. For many practical applications the service area includes goal locations which are both near and far from the start locations, { which necessitates some agents to travel much longer than others, regardless of the goal assignment,} so the advantage is substantial. The results of Section \ref{sec:simulation_results} demonstrate this advantage quantitatively, despite the minor degradation in flight times due to collision detection and resolution, which are discussed next.

\section{Collision detection} \label{sec:collision_detection}
Here the advantage of piecewise polynomial trajectories on straight line paths becomes apparent as the global minimum distance between any pair of agents across their entire trajectories becomes extremely easy and fast to compute. Additionally, the cylindrical collision volume representation synergizes with the restriction that paths are only vertical or horizontal and makes collision checking especially convenient and computationally efficient. Collisions at an instant of time are detected exactly by simply checking if both the radial separation is less than the sum of the radii and the vertical separation is less than the sum of the half-heights. Mathematically, the following equivalent conditions of collision between agents $i$ and $j$ hold:
\begin{numcases}{\mathcal{C}_i \cap \mathcal{C}_j \neq \emptyset \leftrightarrow}
    \|x_{j,12}-x_{i,12}\| \leq R_i+R_j \label{eq:horz_collide} \\
    \land \ |x_{j,3}-x_{i,3}| \leq \frac{H_i+H_j}{2} \label{eq:vert_collide} .
\end{numcases}

For a pair of points moving on straight-line paths whose positions are polynomials in time, the procedure in Alg. \ref{algo:separation_minimization} is used to find the minimum separation distance.
\begin{algorithm} \label{algo:separation_minimization}
\DontPrintSemicolon
 \KwIn{Heading unit vectors $\hat{h}_i$ and $\hat{h}_j$, polynomial trajectories $x_i(t) = p_i(t)\hat{h}_i$ and $x_j(t) = p_j(t)\hat{h}_j$ of degree $d$ over a common time interval $\mathcal{T}_{ij} = [t_0,t_f]$.}
    Calculate relative position polynomial $x_{ij}(t) = p_j(t)\hat{h}_j-p_i(t)\hat{h}_i.$ \;
    Calculate squared separation distance polynomial of degree $2d+1$ as $p_{ij}(t) = x_{ij}(t)^\intercal x_{ij}(t)$ whose coefficients are computed from multiplication and addition of the appropriate coefficients of $x_{ij}(t)$.\;
    Minimize the squared separation distance using Algorithm \ref{algo:polynomial_minimization} with inputs $p_{ij}(t)$ and $[t_0,t_f]$.\;
 \KwOut{Minimum separation distance $d^* = \sqrt{\underset{t \in [t_0,t_f]}{\min} x_{ij}(t)^\intercal x_{ij}(t)}$.}
 \caption{Separation minimization}
\end{algorithm}

Consequently, Alg. \ref{algo:segment_pair_collision} is used to check for a collision between a pair of agents for a single pair of polynomial segment trajectories. The algorithm uses a default return of ``false'' and terminates immediately whenever ``true'' is returned; this ``short-circuiting'' dramatically improves computational speed. First, it is checked whether the intersection of the time intervals for the segments is nonempty; otherwise the segments are never active at the same time and no collision could occur. 
Then it is determined whether both agents are moving vertically, both horizontally, or one of each. Based on this, it is checked if \eqref{eq:horz_collide} or \eqref{eq:vert_collide} are satisfied, and if so Alg. \ref{algo:separation_minimization} is used to obtain the relevant minimum separation distance and that distance is used in \eqref{eq:horz_collide} or \eqref{eq:vert_collide} to determine the presence of a collision.

Now segment pair collision detection is used to detect collisions between all pairs of full composite trajectories of the polynomial segment type described earlier using Alg. \ref{algo:all_agent_collision}. The paths start and end at the start and goal locations on the ground and reside entirely within the planar region with infinite vertical extent passing through the line segment joining the start and goal i.e. $p_i(t) \in \{x | x_{12} \in \ell_i \} \forall t$. A ``short-circuit" of the full polynomial segment collision check is then accomplished by first doing a computationally cheap check which helps quickly guarantee safety of many trajectory segment pairs. If the minimum distance between the two line segments of the trajectory pair is greater than the sum of agent radii, then a collision is impossible since there is no configuration of the agent centers within the assumed planar regions which gives an intersection since \eqref{eq:horz_collide} is impossible to satisfy. If the minimum distance between the two line segments of the trajectory pair is not greater than the sum of agent radii, then  collision detection using Alg. \ref{algo:segment_pair_collision} is run. Using this fast preliminary check is critical to obtaining usable performance since, in all but the most highly congested scenarios, this check catches a large portion of segment pairs which are far apart spatially.
The segment collision check is repeated for all polynomial trajectory segment pairs (for a single pair of agents).
As soon as a collision is detected on a single pair of trajectory segments, the pair of agents is flagged as having a collision and the check progresses to the next pair of agents without finishing checking all remaining segments of the current pair of agents (another ``short-circuit'').
This process is repeated for each pair of agents, resulting in a symmetric boolean matrix of collision flags which can be represented by an upper triangular matrix or flattened vector to reduce the storage space by half.
With exact collision results for the entire group of agents and trajectories in hand, the proposed methodology advances on to resolving the detected collisions.

\begin{algorithm} \label{algo:all_agent_collision}
\DontPrintSemicolon
 \KwIn{Collection of $n$ trajectories $\gamma_k(t)$ for $k=1,\ldots,n$.}
    \ForEach{Pair of agents $i,j$}{
    Calculate the minimum distance $\delta_{ij}^*$ between the two line segments joining the starts and goals e.g. via \cite{Lumelsky1985}. \;
    \eIf{$\delta_{ij}^* > R_i+R_j$}{
        $B_{ij} \gets$ False
    }{
        \ForEach{Pair of segments $\gamma_{im}$ in $\gamma_i(t)$ and $\gamma_{jn}$ in $\gamma_j(t)$}{
            $B_{ij} \gets$ result of Alg. \ref{algo:segment_pair_collision} with inputs $\gamma_{im}$ and $\gamma_{jn}$. \;
            \If{$B_{ij} = $ True}{
                \Break
            }
        }    
    }
}
 \KwOut{Boolean matrix $F \in \mathbb{S}^{n \times n}$ of collision flags.}
 \caption{All agents collision check}
\end{algorithm}

\section{Collision resolution}

The overall trajectory generation proceeds by using the general collision detection scheme described in the previous section to determine which agents collide assuming they are all in the same altitude.
After single-altitude collisions are detected, they are resolved by inserting vertical trajectories and time delays and/or additional altitudes.

\subsection{Collision resolution via time delay} \label{sec:collision_resolution_time}
\begin{figure}[pos=h] 
    \centering
  \subfloat[\label{1a}]{%
      \includegraphics[width=1.6in]{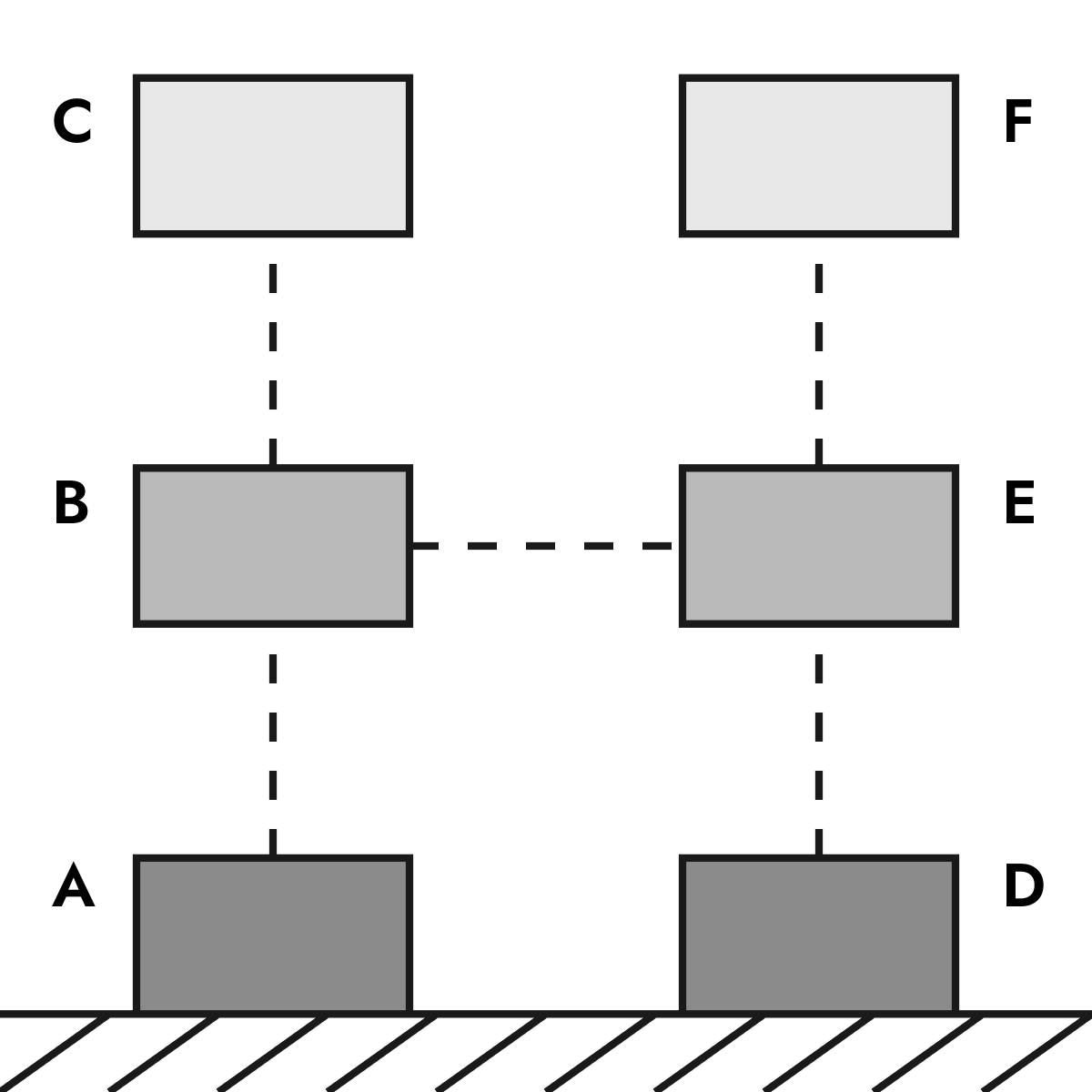}}
    \hfill
  \subfloat[\label{1b}]{%
        \begin{tabular}[b]{||c c c||} 
         \hline
          $t$ & $x_1$ & $x_2$ \\
         \hline\hline
         1 & A & D \\ 
         \hline
         2 & C & F \\
         \hline
         3 & B & F \\
         \hline
         4 & E & F \\
         \hline
         5 & D & F \\
         \hline
         6 & D & E \\
         \hline
         7 & D & B \\
         \hline
         8 & D & A \\
         \hline
        \end{tabular}
        }
  \caption{(a) Diagram and (b) table for a case which motivates the ability of time delay to resolve collisions.}
  \label{fig:figure_delay_times1}
\end{figure}

One way to resolve collisions is to send all agents first to a high holding altitude $\mathcal{A}_{\text{hold}}$, then after some delay times have agents descend vertically down, then move horizontally in a single traversal altitude $\mathcal{A}_{\text{trav}}$, then finally descend to the ground altitude $\mathcal{A}_{\text{gnd}}$ at the goal location. In this scheme, a maximum of three altitudes are needed with a total height of $2.5H$ above the ground plane. By construction, given sufficient delay time on each agent that eventually all agents can complete their trajectories without colliding, since in the worst case an agent can simply wait in the holding altitude until all other agents have completed their trajectories and landed. See Fig. \ref{fig:figure_delay_times1} for an illustration of this idea in the case when two identical agents must exchange positions. Although such a troublesome goal assignment would never be chosen by the goal assignment procedure in Sec. \ref{sec:goal_assignment} since the reversal of the assignment gives a lower cost, it is conceptually useful simply to illustrate the ability of time delay to resolve collisions. The image shows a side-view with dashed lines representing paths and the table shows a sequence of positions that the agents pass through at generalized times $t$ along the linear paths.

The reason to have agents wait in a high holding altitude rather than on the ground is simply that the start and goal locations of two agents may be within a colliding distance of eachother. If the somewhat weak restriction is imposed that 
\begin{align}
    \|s_i-g_j\| \geq 2 R \ \forall \ i \neq j \in \mathcal{I}_n
\end{align}
then the possibility of landing on top of another agent waiting on the ground is avoided and the holding altitude is unnecessary and agents can wait on the ground i.e. set $\eta_{\text{hold}}=0$ so that the holding and ground altitudes coincide. In either case, the proposed method works the same way. 

Choosing the height of the traversal and holding altitudes as $\eta_{\text{trav}} = \eta_{\text{gnd}} + H = H$ and $\eta_{\text{hold}} = \eta_{\text{trav}} + H = 2H$ ensures agents in different altitudes cannot collide. The traversal and holding altitudes are thus $\mathcal{A}_{\text{trav}} = \mathcal{A}(\eta_{\text{trav}})$ and $\mathcal{A}_{\text{hold}} = \mathcal{A}(\eta_{\text{hold}})$.

Each full trajectory is made up of 4 or 5 subtrajectories which are generated according to the procedure in Sec. \ref{sec:traj} which have 2 or 3 polynomial segments each:
\begin{enumerate}
    \item Vertical ascent from $\mathcal{A}_{\text{gnd}} \rightarrow \mathcal{A}_{\text{hold}}$: 
        \begin{align} \label{eq:time_delay_segment_1} s_i \rightarrow s_i + [0,0,\eta_{\text{hold}}]^\intercal \end{align}
    \item Stationary wait in $\mathcal{A}_{\text{hold}}$ for time $\tau_i$: 
        \begin{align} \label{eq:time_delay_segment_2} s_i + [0,0,\eta_{\text{hold}}]^\intercal \rightarrow s_i + [0,0,\eta_{\text{hold}}]^\intercal \end{align} 
    \item Vertical descent from $\mathcal{A}_{\text{hold}} \rightarrow \mathcal{A}_{\text{trav}}$: 
        \begin{align} \label{eq:time_delay_segment_3} s_i + [0,0,\eta_{\text{hold}}]^\intercal  \rightarrow s_i + [0,0,\eta_{\text{trav}}]^\intercal \end{align}
    \item Horizontal movement within $\mathcal{A}_{\text{trav}}$: 
        \begin{align} \label{eq:time_delay_segment_4} s_i + [0,0,\eta_{\text{trav}}]^\intercal  \rightarrow g_i + [0,0,\eta_{\text{trav}}]^\intercal \end{align}
    \item Vertical descent from $\mathcal{A}_{\text{trav}} \rightarrow \mathcal{A}_{\text{gnd}}$: 
        \begin{align} \label{eq:time_delay_segment_5} g_i + [0,0,\eta_{\text{trav}}]^\intercal \rightarrow g_i \end{align}
\end{enumerate}
In the case that agents wait on the ground, the subtrajectory in step 1 can be skipped and the agents ascend rather than descend in step 3.

The trajectory generation problem is now reduced to finding the set of time delays $\tau_i$ whose sum is minimum and also resolve all collisions while adhering to the trajectory generation framework described earlier:
\begin{align} \label{eq:delay_opt_problem}
    \underset{\tau}{\text{minimize  }} &\sum_{i=1}^{n} \tau_i \\
    \text{subject to  } & \eqref{eq:collision_free_constraint}, \gamma(t) \leftarrow [\eqref{eq:time_delay_segment_1}, \eqref{eq:time_delay_segment_2}, \eqref{eq:time_delay_segment_3}, \eqref{eq:time_delay_segment_4}, \eqref{eq:time_delay_segment_5}] \nonumber 
\end{align}
where \eqref{eq:collision_free_constraint} is the boolean collision avoidance constraint whose value is determined by the collision detection scheme in Sec. \ref{sec:collision_detection}.
This problem is nonconvex due to the collision avoidance constraint and has continuous decision variables, so a discretization scheme is used as an effective heuristic. The heuristic begins by ordering the agents randomly, then for each agent the associated delay time is increased by an increment $\tau_\Delta$ until collisions with all agents whose time delays have been fixed are resolved. This is repeated until each agent's delay time has been established. This procedure is expressed in Alg. \ref{algo:time_delays}.

\begin{algorithm} \label{algo:time_delays}
\DontPrintSemicolon
 \KwIn{Collection of $n$ trajectories $\gamma_k(t)$ for $k=1,\ldots,n$.}
    \For{$i = 1, \ldots, n$ }{
    Initialize $\tau_i=0$ \;
    \While{Any $F \leftarrow$ Alg. \ref{algo:all_agent_collision} ($\gamma_k(t)$ for $k=1,\ldots,i$)}{
    $\tau_i \leftarrow \tau_i + \tau_\Delta$ \;
    Apply time delay $\tau_i$ to trajectory $\gamma_i(t)$ }
}
 \KwOut{Collection of $n$ collision-free trajectories $\gamma_k(t)$ with included time delays $\tau_k$ for $k=1,\ldots,n$}
 \caption{Collision resolution via time delays}
\end{algorithm}

Although this heuristic is not assured to find the global minimum of the problem in \eqref{eq:delay_opt_problem}, the solutions found are empirically good and importantly are guaranteed to be found after a finite number of computations. To see this, consider the case depicted in Fig. \ref{fig:figure_delay_times1} where each agent proceeds one-by-one; the first agent descends from the holding altitude and completes its full trajectory while all remaining agents remain in the holding altitude, then the second agent does the same and so forth until all agents have landed. 

Let $T_j$ for $j=1,\ldots,n$ be the times taken by each agent to execute trajectory segments 3, 4, 5 in \eqref{eq:time_delay_segment_3}, \eqref{eq:time_delay_segment_4}, \eqref{eq:time_delay_segment_5}, with $T_j$ ordered from greatest to least. Then an upper bound on the number of increments of time delay increase for any other agent is $n_{\text{inc}, \max} \leq n_{\text{inc}, \max} = \text{ceil}(T_1/\tau_\Delta)$ since for any greater time delay a collision is not possible, as explained earlier in the discussion of Fig. \ref{fig:figure_delay_times1}. Applying this argument iteratively shows that an upper bound on the number of increments for $\tau_i$ is $n_{\text{inc}, i} \leq i \times n_{\text{inc}, \max}$, and thus an upper bound on the total number of increments is 
\begin{align}
    n_{\text{inc}, \text{tot}} = \sum_{i=1}^n n_{\text{inc}, i} \leq \frac{n(n+1)}{2} n_{\text{inc}, \max} ,
\end{align}
which is clearly $\mathcal{O}(n^2)$.
In practice, many fewer increments are required than this conservative upper bound.
The ordering of agents could likely be further improved i.e. according to some metric such as shortest time in horizontal flight, but it was found that random ordering gave good results.

\subsection{Collision resolution via altitude assignment} \label{sec:collision_resolution_altitude}
Another way to resolve collisions is by finding an assignment to a set of altitudes and sending agents on trajectories that move horizontally only in these altitudes. The altitudes are given sufficient vertical separation to ensure clearance between agents in different altitudes regardless of horizontal position. Additional wait time and holding altitudes are introduced to resolve potential secondary collisions induced by the primary collision resolution. 

There are $m$ traversal altitudes $\mathcal{A}_{\text{trav,i}}$ for $i \in \mathcal{I}_m$ and $h$ holding altitudes $\mathcal{A}_{\text{hold,i}}$ which are inserted between traversal altitudes and indexed to match the traversal altitudes, although in general $h \leq m-1$.
In this scheme, a maximum of $n$ traversal altitudes and $n$ holding altitudes are needed in addition to the ground altitude.

Define the $n \times m$ boolean altitude assignment matrix $B$, which assigns agents to altitudes, as
\begin{equation} 
B_{ij} = 
\begin{cases}
1 & \text{if agent } i \text{ is assigned to altitude } j\\
0 & \text{otherwise}
\end{cases}
\end{equation}
Therefore in row $i$ of $B$, denoted as $B_i$, the index where $B_{ij}=1$ gives the altitude assigned to agent $i$. Alternatively, in column $j$ of $B$ the indices where $B_{ij}=1$ give the agents assigned to altitude $j$.
All agents are assigned to altitudes in a one-to-many mapping, so
\begin{equation} \label{eq:assignment_alts}
B^\intercal B = D_{m}
\end{equation}
where $D_m$ is an $m\times m$ diagonal matrix whose entry $D_{ii}$ is the integer number of agents assigned to altitude $i$.

Each full trajectory is made up of 4 or 6 subtrajectories which are generated according to the procedure in Sec. \ref{sec:traj} which have 2 or 3 polynomial segments each:
\begin{enumerate}
    \item Vertical ascent from $\mathcal{A}_{\text{gnd}} \rightarrow \mathcal{A}_{\text{trav,i}}$: 
        \begin{align} \label{eq:altitude_segment_1} s_i \rightarrow s_i + [0,0,\eta_{\text{trav,i}}]^\intercal \end{align}
    \item Stationary wait in $\mathcal{A}_{\text{trav,i}}$ until global time $t_1$: 
        \begin{align} \label{eq:altitude_segment_2} s_i + [0,0,\eta_{\text{trav,i}}]^\intercal \rightarrow s_i + [0,0,\eta_{\text{trav,i}}]^\intercal \end{align} 
    \item Horizontal movement within $\mathcal{A}_{\text{trav,i}}$: 
        \begin{align} \label{eq:altitude_segment_3} s_i + [0,0,\eta_{\text{trav,i}}]^\intercal  \rightarrow g_i + [0,0,\eta_{\text{trav,i}}]^\intercal \end{align}    
    \item Vertical descent from $\mathcal{A}_{\text{trav},i} \rightarrow \mathcal{A}_{\text{hold},i}$: 
        \begin{align} \label{eq:altitude_segment_4} g_i + [0,0,\eta_{\text{trav},i}]^\intercal  \rightarrow s_i + [0,0,\eta_{\text{hold},i}]^\intercal \end{align}
    \item Stationary wait in $\mathcal{A}_{\text{hold,i}}$ for time $\tau_i$: 
        \begin{align} \label{eq:altitude_segment_5} s_i + [0,0,\eta_{\text{hold,i}}]^\intercal \rightarrow s_i + [0,0,\eta_{\text{hold,i}}]^\intercal \end{align} 
    \item Vertical descent from $\mathcal{A}_{\text{hold},i} \rightarrow \mathcal{A}_{\text{gnd}}$: 
        \begin{align} \label{eq:altitude_segment_6} g_i + [0,0,\eta_{\text{hold},i}]^\intercal \rightarrow g_i \end{align}
\end{enumerate}
where the final 3 subtrajectories may be collapsed to a single vertical descent from $\mathcal{A}_{\text{trav},i} \rightarrow \mathcal{A}_{\text{gnd}}$.

Similarly to the collision resolution via time delays, the trajectory generation problem is now reduced to finding the altitude assignment and set of time delays $\tau_i$ which minimize the sum of flight times and also resolve all collisions while adhering to the trajectory generation framework described earlier:
\begin{align}
    \underset{B, \tau}{\text{minimize  }} &\sum_{i=1}^{n} \tau_i \\
    \text{subject to  } & \eqref{eq:collision_free_constraint}, \gamma(t) \leftarrow [\eqref{eq:altitude_segment_1}, \eqref{eq:altitude_segment_2}, \eqref{eq:altitude_segment_3}, \eqref{eq:altitude_segment_4}, \eqref{eq:altitude_segment_5}, \eqref{eq:altitude_segment_6}] \nonumber 
\end{align}

\subsubsection{Primary collisions}
Primary collisions are those resulting from two agents moving horizontally in a shared altitude. These are resolved by altitude assignment. As a heuristic for minimizing the sum of flight times, one might seek to minimize the number of altitudes required so that time spent in vertical motion is minimized. However even finding the optimal altitude assignment which minimizes the number of altitudes is a hard nonconvex combinatorial problem, so a similar procedure as in collision resolution via time delays is used to find the altitude assignment $B$. Agents are prioritized randomly, then each agent is assigned the lowest altitude possible that resolves primary collisions with all previously assigned agents. If no such altitude exists, a new one is created at a height above the previous highest altitude by a vertical spacing of $H$. This is repeated for all agents. By construction, such an assignment guarantees that there will be no collisions during the horizontal movements. Alg. \ref{algo:altitude_assignment} documents this procedure using mathematical notation.
\begin{algorithm} \label{algo:altitude_assignment}
\DontPrintSemicolon
 \KwIn{Boolean collision flag matrix $F \in \mathbb{S}^{n \times n}$.}
    Initialize $m=1$ \;
    \For{$i = 1, \ldots, n$ }{
        \For{$j = 1, \ldots, m$ }{
            \uIf{not any $F_{i,k}$ for $k \ | \ B_{k,j}==\text{True}$}{
                $B(i, j) = $ True \;
            }
            \uElseIf{$j==m$}{
                $m \leftarrow m+1$ \;
                $B(i, m) = $ True \;
            }
            \Else{
                Continue
            }
        }
    }
 \KwOut{Boolean altitude assignment matrix $B \in \mathbb{R}^{n \times m}$ that resolves primary collisions.}
 \caption{Altitude assignment}
\end{algorithm}

\subsubsection{Secondary collisions}
Although altitude assignment resolves primary collisions, the possibility remains of secondary collisions during the vertical descent movements down towards the goals on the ground. These are easily detected by the same collision detection scheme in Sec. \ref{sec:collision_detection}. Secondary collisions are exhaustively partitioned into two types of collision: exit and entrance collisions. In practice, it was found that these secondary collisions were exceedingly rare, but nevertheless must be prevented.

\paragraph{Exit collisions}
In an exit collision, a descending agent is struck by another agent moving horizontally in the same traversal altitude. To resolve this, a simple enlargement of the collision radius is used. The longest time $T_\text{exit}$ that any agent could take to exit its altitude is calculated; this is easily accomplished by generating a trajectory which descends vertically downwards by $H$ (the spacing between two altitudes). This captures the effect of all position derivative constraints imposed on the agents. This is also conservative since some agents may not have to come to a full stop at the altitude below; some agents will continue descending and accelerating which would reduce the time taken to exit the altitude, but this is ignored for simplicity. Next, the greatest distance $L_\text{exit}$ that the fastest agent would traverse horizontally moving at maximum speed over the time $T_\text{exit}$ is calculated. Then the collision radii of all agents are increased by $L_\text{exit}/2$. Thus by using the same collision detection scheme in Sec. \ref{sec:collision_detection} it is ensured that agents maintain an additional horizontal clearance of $L_\text{exit}$ at all times, which by construction ensures that exit collisions are impossible.

Unfortunately this procedure requires the collision radii to be increased by an amount proportional to the maximum speed of the agents, but for agents with high maximum acceleration relative to the maximum speed, such as quadrotors, the detriment is not too severe. The enlargement of the collision radii is performed as the first step of the overall collision resolution, prior to finding the altitude assignment to resolve primary collisions.

\paragraph{Entrance collisions}
In an entrance collision, a descending agent enters a lower traversal altitude at the same time as another agent is moving horizontally underneath. To resolve an entrance collision, a holding altitude is placed between the descending agent's traversal altitude and the next lowest traversal altitude (if one does not already exist). This gives the descending agent a place to wait while the other agent moves out of the way. Once the holding altitude has been inserted, new trajectories are generated and the entire check must begin again from the point where the altitude assignment was made. In particular, the offending descending agent is made to come to a full stop and wait in its (newly inserted) holding altitude. If an entrance collision still exists with this agent, delay time is added according to the same scheme as in Section \ref{sec:collision_resolution_time}. Again, by construction, given sufficient delay time all the possible collisions with agents at lower heights will be resolved since those agents can all land. Agents in the lowest traversal altitude will clearly not encounter this type of secondary collision, and so can complete their trajectories without collision. Arguing inductively, since the lowest agents have collision-free trajectories, and entrance collisions can be resolved for agents in each successively higher altitude, all entrance collisions can be resolved. Since there are a finite number of altitudes, it also follows that the time delays required are also finite.
The roles of each agent in an entrance collision are distinguished by the collision detection algorithm simply by noting the heading vector of each agent.
{ As a final remark,} in the worst case $n$ traversal altitudes and $n$ holding altitudes are needed, and thus by construction the computations terminate in finite time.

\begin{figure}[pos=h]
\centering
\includegraphics[width=2.5in]{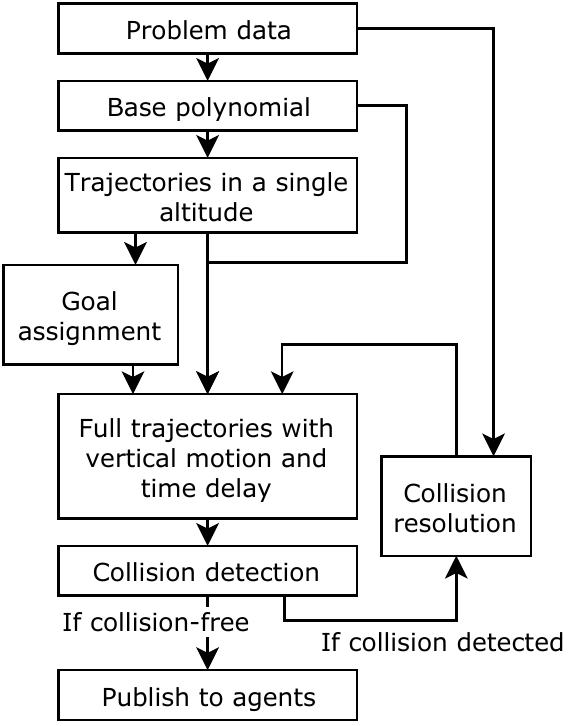}
\caption{Diagram of information flow for trajectory generation.}
\label{fig:flow}
\end{figure}

{
To conclude the algorithmic development, Figure \ref{fig:flow} gives a broad description of all the steps involved in the method and their relationships.
Evaluation of the proposed schemes is presented next, both in computer simulations and in deployment on a physical testbed.
}

\section{Simulation results} \label{sec:simulation_results}
Both the computer simulations and physical experiments were based on the Crazyswarm, a hardware and software platform that serves as a research testbed for quadrotor autonomy, which is described in more detail in Section \ref{sec:experimental_results}.
Throughout the simulations, the vehicle parameters used for trajectory planning were chosen to match the actual Crazyswarm platform used in the physical experiments. 
The Crazyflie quadrotor vehicles had a nominal outer diameter of 14 cm and height of 4 cm, while the diameter and height of the cylinders used for trajectory planning were enlarged to 30cm and 40cm respectively; see Section \ref{sec:experimental_results} for the rationale of this enlargement.
The kinematic constraints imposed during trajectory generation for all simulations are listed in Table \ref{tab:table_kinematic_constraints}. These correspond to conservative values computed by scaling down the most aggressive values the physical Crazyswarm platform could experience without significant tracking error.

\begin{table} \label{tab:table_kinematic_constraints}
\caption{Kinematic constraints}
\centering
\normalsize
\begin{tabular}{ c c c }
\toprule
\ & \multicolumn{2}{c}{Upper limit}\\
Time derivative & Horizontal & Vertical \\
\midrule\\
\addlinespace[-3ex]
Speed $(m/s)$ & 0.2 & 0.2
\\
Acceleration $(m/{s^2})$ & 0.5 & 0.5
\\
Jerk $(m/{s^3})$ & 10 & 10
\\
\bottomrule
\end{tabular}
\end{table}
For the time delay increase rule, used by both algorithms discussed in Sections \ref{sec:collision_resolution_time} and \ref{sec:collision_resolution_altitude}, an addition rule with an increment of $\tau_\Delta = 0.1 s$ was used, which was found empirically to strike a nice balance between computation time and quality of solutions.

In order to analyze the performance of the proposed algorithms, Monte Carlo trials were performed with start and goal locations generated randomly with uniform probability over a square of side length $S$. In all trials all agents were identical so that collision volume dimensions were $R_i=R$, $H_i=H$ and position derivatives were $\delta_i=\delta$ for all $i \in \mathcal{I}_n$. The number of agents $n$ and the area density $\eta$ were varied, where $\eta$ is defined as the ratio of the summed area of all agents' projection onto the ground to the area on the ground that any projection could occupy:
\begin{equation*}
\eta = \frac {A_{\text{agents}}}{A_{\text{space}}} = \frac{n\pi R^2}{S^2+4RS+\pi R^2}.
\end{equation*}
To ensure initial and terminal configurations were noncolliding, a minimum start-start and goal-goal separation distance of $2R$ was imposed. This led to an upper bound on the density, which occurs when the start locations are hexagonally close packed; proofs of this fact date back to Lagrange in 1773 with the first universally accepted proof delivered by Toth in 1942 \cite{Toth1942}. For a separation of $2R$ the upper limit of density is $\frac{\pi}{2\sqrt{3}}\approx0.9068$ and for a separation of $2\sqrt{2}R$ as in~\cite{turpin2014} the limit is $\frac{\pi}{4\sqrt{3}}\approx0.4534$. For reference, a typical area density encountered in commercial aircraft traffic management is on the order of $10^{-5}$~\cite{air2016}. For applications involving many unmanned aerial robots the traffic is considerably more dense, so simulations were performed over a wide range of densities.

\subsection{Time delay distribution}
\begin{figure}[pos=h]
\centering
\includegraphics[width=3in]{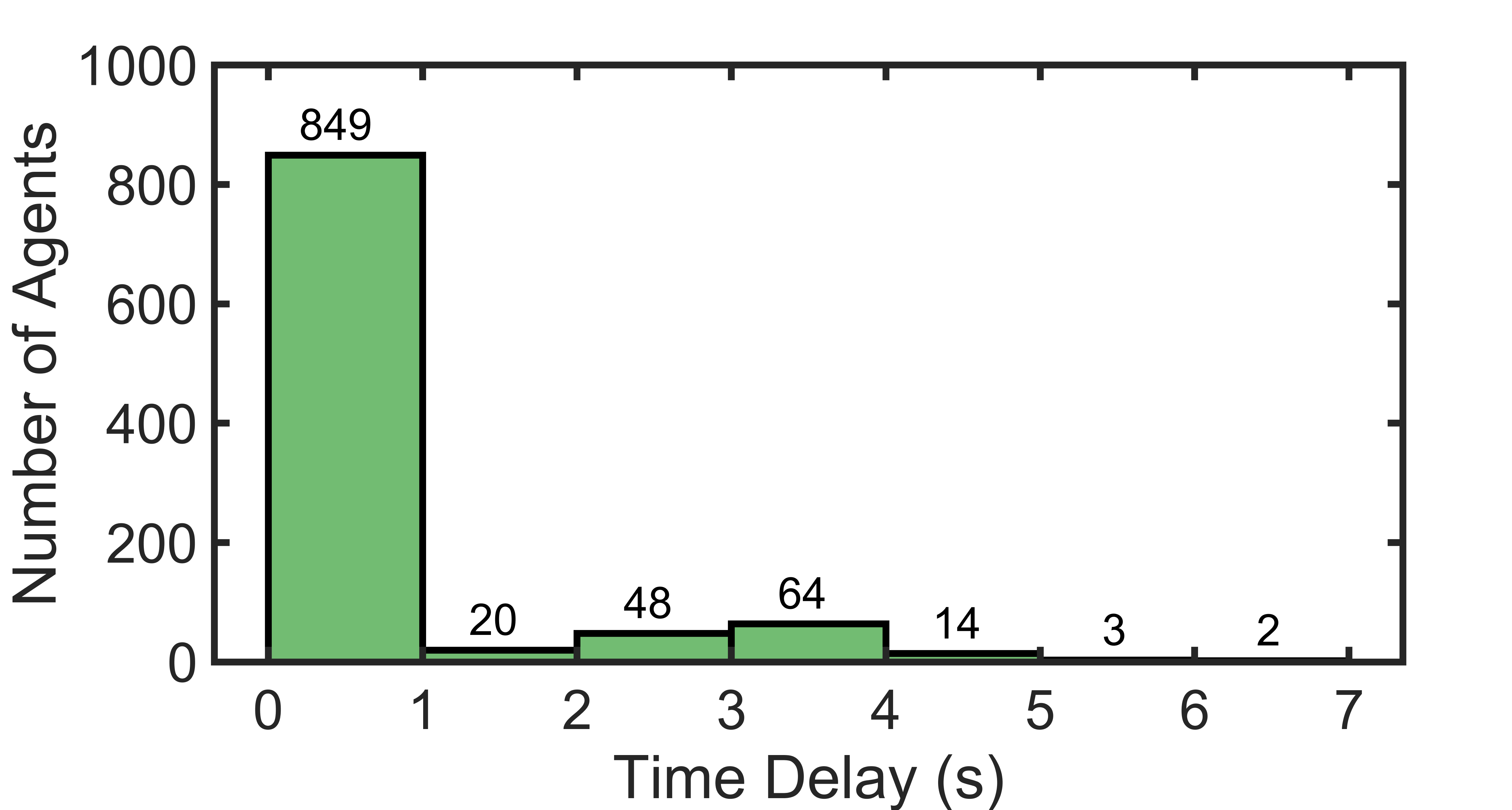}
\caption{Histogram of time delays for 1000 agents with $\eta = 10^{-1/2} \approx 0.31$.}
\label{time_delay_histogram}
\end{figure}

Fig.~\ref{time_delay_histogram} shows a histogram of time delays using the proposed time delay collision resolution method for $n=1000$ agents and a high density of $\eta = 10^{-1/2} \approx 0.31$ { for a single random Monte Carlo problem instance described in Section \ref{sec:simulation_results}}. This shows that, even when start and goal locations are very dense, most agents have zero or low-valued time delays.

\subsection{Number of altitudes required}
\begin{figure}[pos=h]
    \centering
    \includegraphics[width=2in]{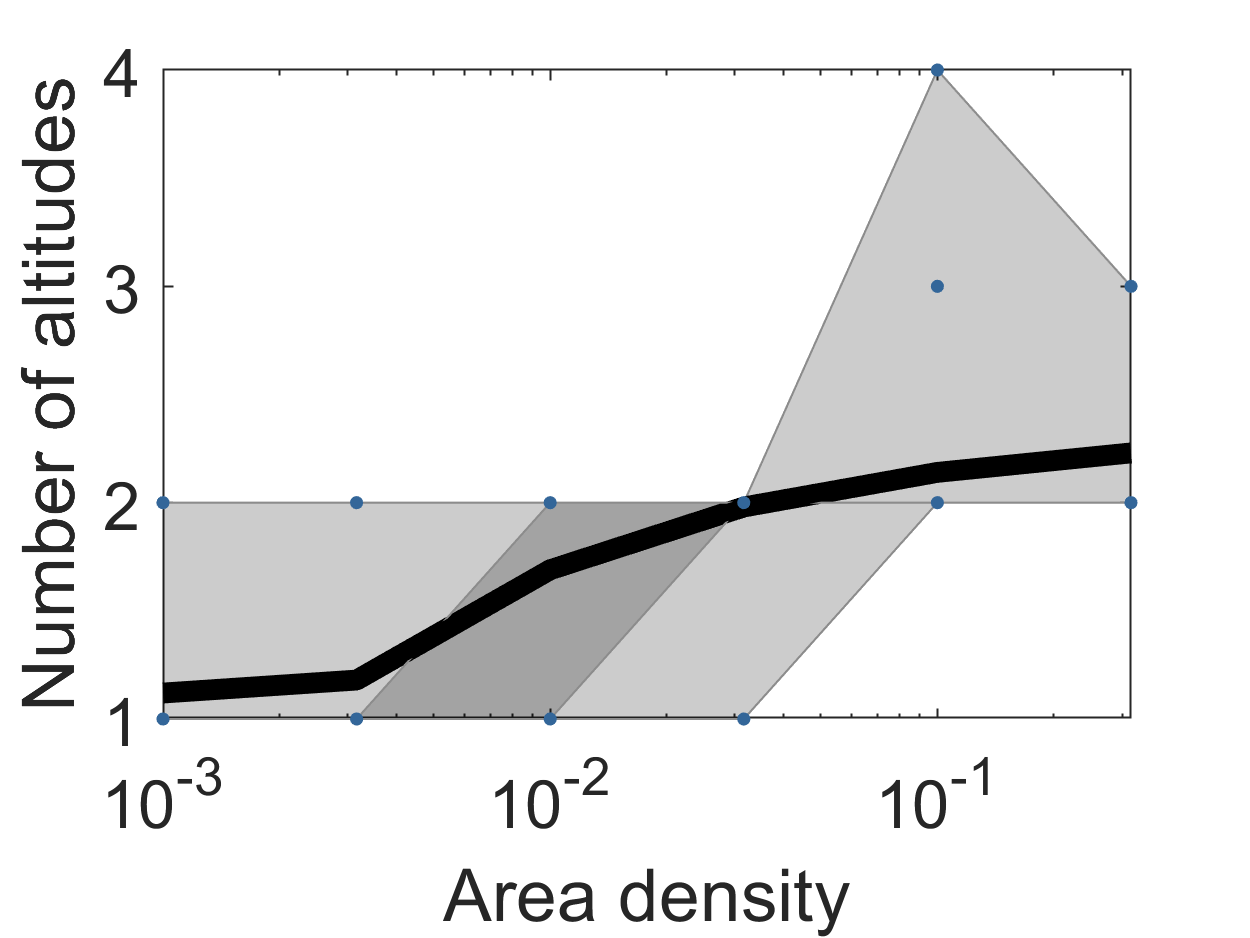}
    \caption{Number of flight altitudes required to resolve collisions as a function of area density. The minimum possible number of altitudes was 1. The number of agents was held constant at $n = 100$ and 100 trials were run at each density. Individual data for each agent in each trial are plotted as points. The mean, interquartile range (25th to 75th percentile), and full range (0th to 100th percentile) are plotted as a bold black line, dark shaded region, and light shaded region. The vertical axis is linear scaled and the horizontal axis is log scaled with base 10. }
    \label{fig:AltQuantMonteCarlo}  
\end{figure}

Using altitude assignment, the number of altitudes required to resolve collisions was studied. Fig.~\ref{fig:AltQuantMonteCarlo} shows the number of flight altitudes (altitudes other than the ground) as a function of area density. As expected, the number of altitudes required grew as the density increased as a result of more potential collisions, but only a few altitudes were required even for highly dense scenarios.

\subsection{Normalized flight times}
To analyze the relative degradation in flight times due to avoiding collisions (altitude changes and time delays), the flight time data are normalized by dividing by the average time spent in horizontal motion for each trial.
The time spent in horizontal motion can be viewed as unavoidable, since this is the minimum time which must be spent to reach the goals even if collisions were ignored.

\subsubsection{Effect of collision resolution}
\begin{figure}[pos=h] 
\centering
  \subfloat[Horizontal motion time\label{fig:monteCarloDelays_a}]{%
      \includegraphics[width=1.6in]{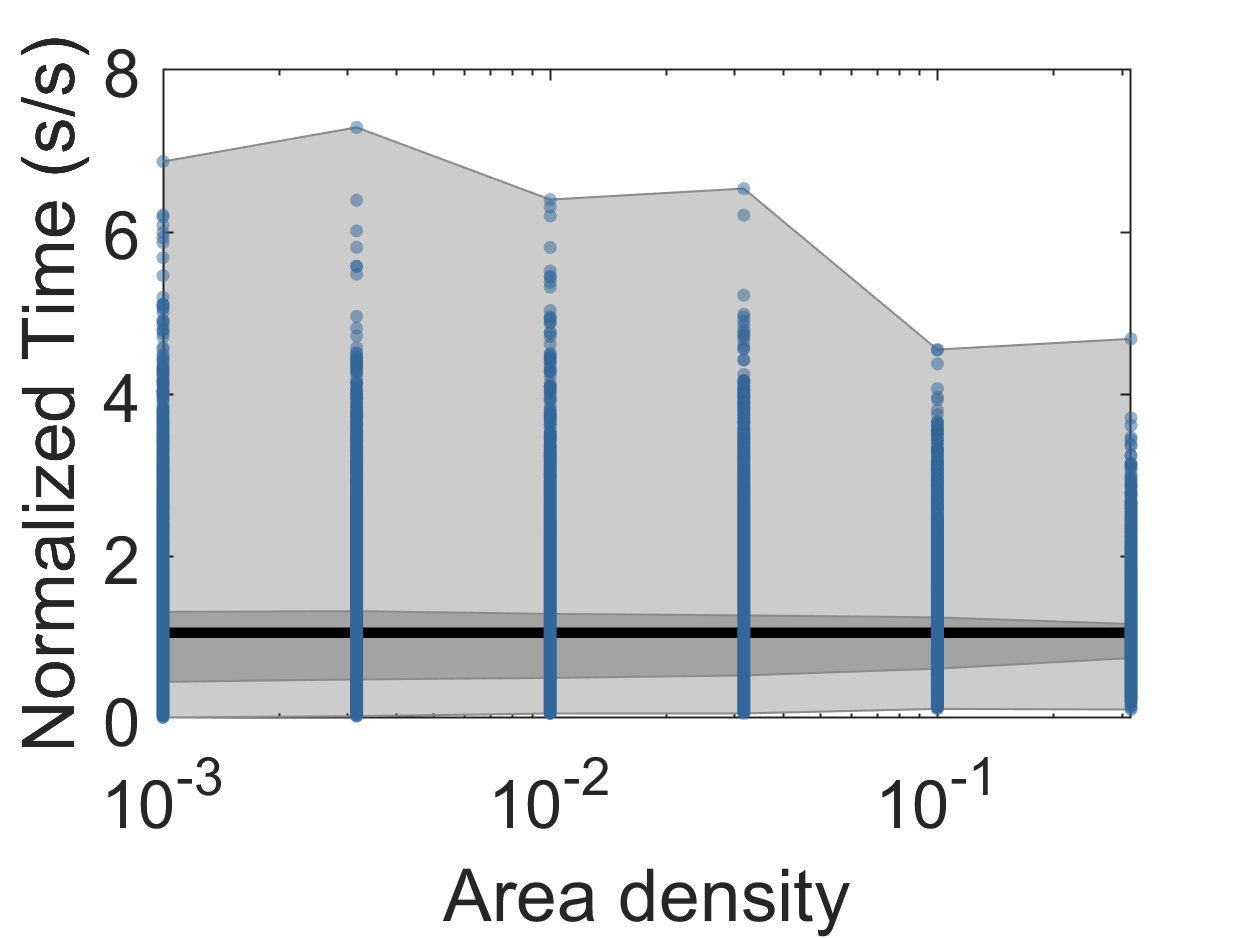}}
    \hfill
  \subfloat[Vertical motion time\label{fig:monteCarloDelays_b}]{%
      \includegraphics[width=1.6in]{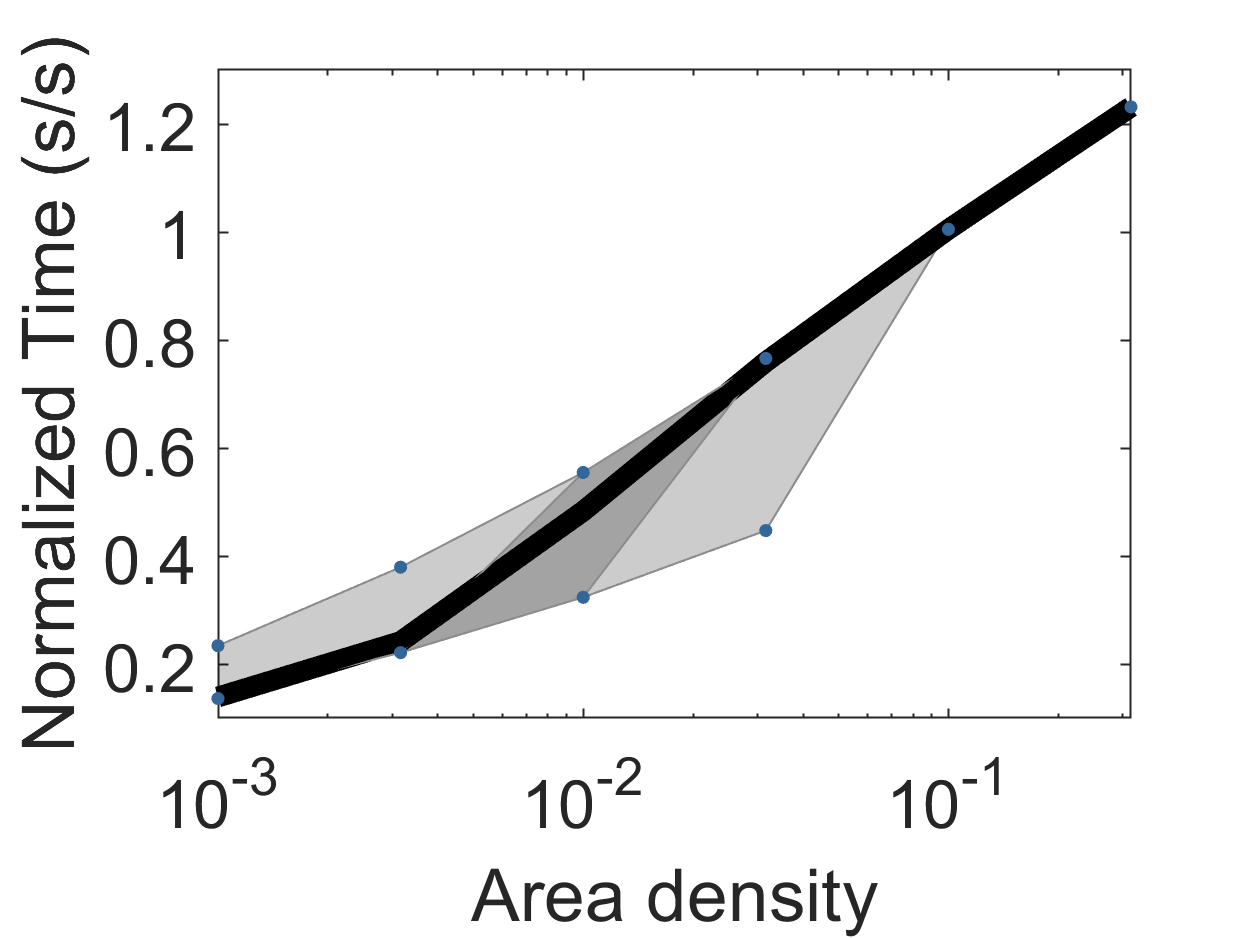}}  
    \hfill
  \subfloat[Waiting time\label{fig:monteCarloDelays_c}]{%
      \includegraphics[width=1.6in]{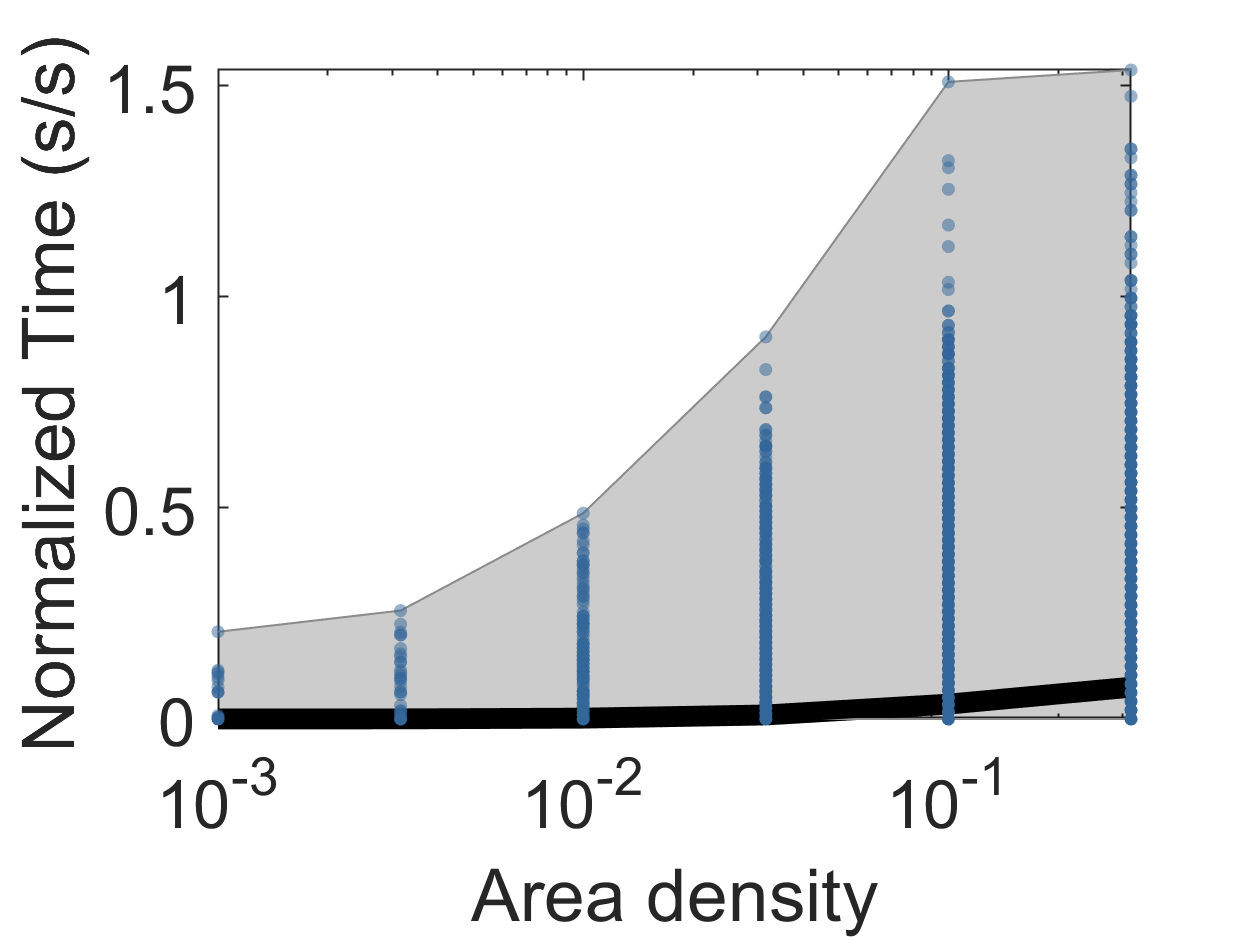}}
    \hfill
  \subfloat[Total time \label{fig:monteCarloDelays_d}]{%
      \includegraphics[width=1.6in]{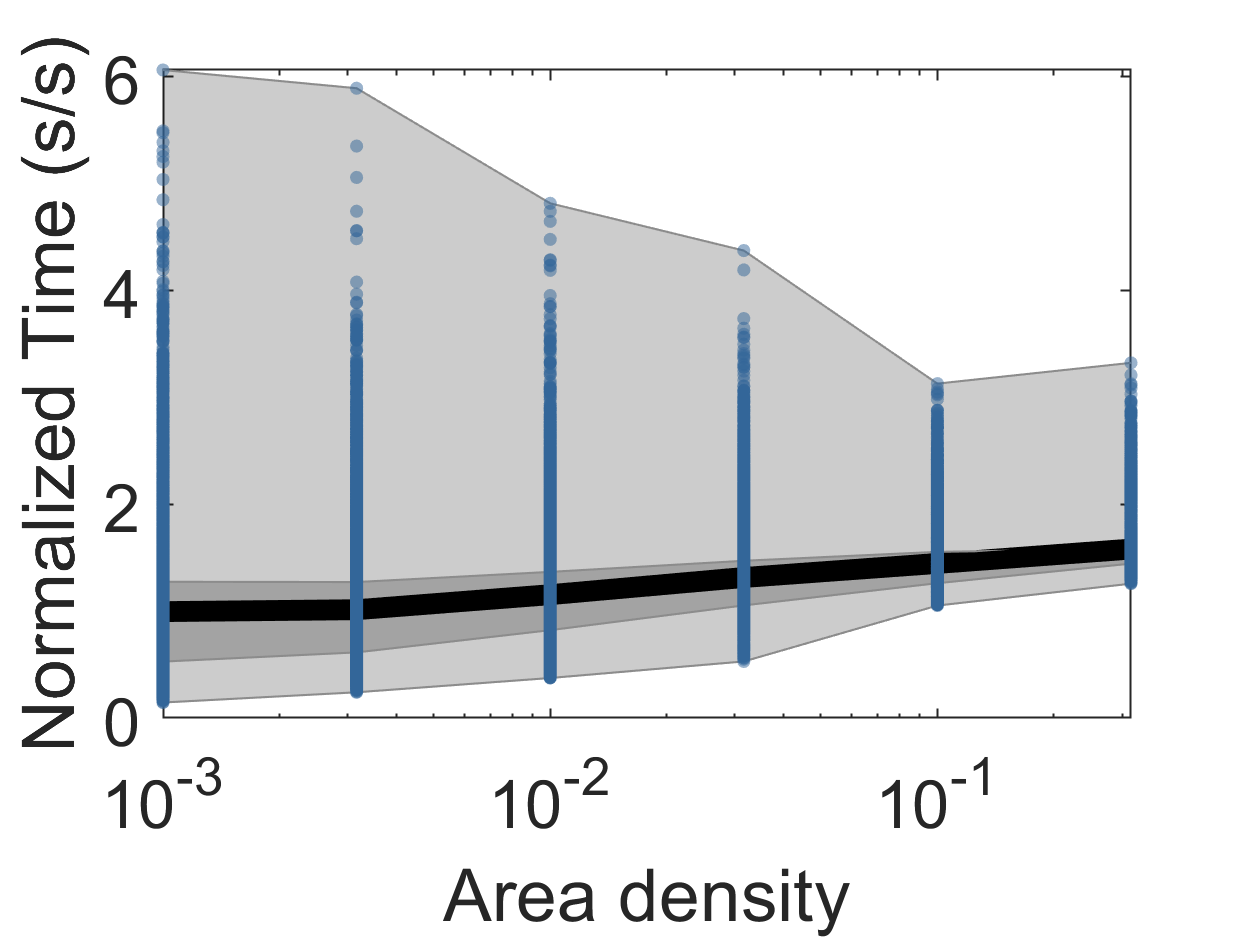}}
    \hfill
  \subfloat[Total time, zoomed \label{fig:monteCarloDelays_e}]{%
      \includegraphics[width=1.6in]{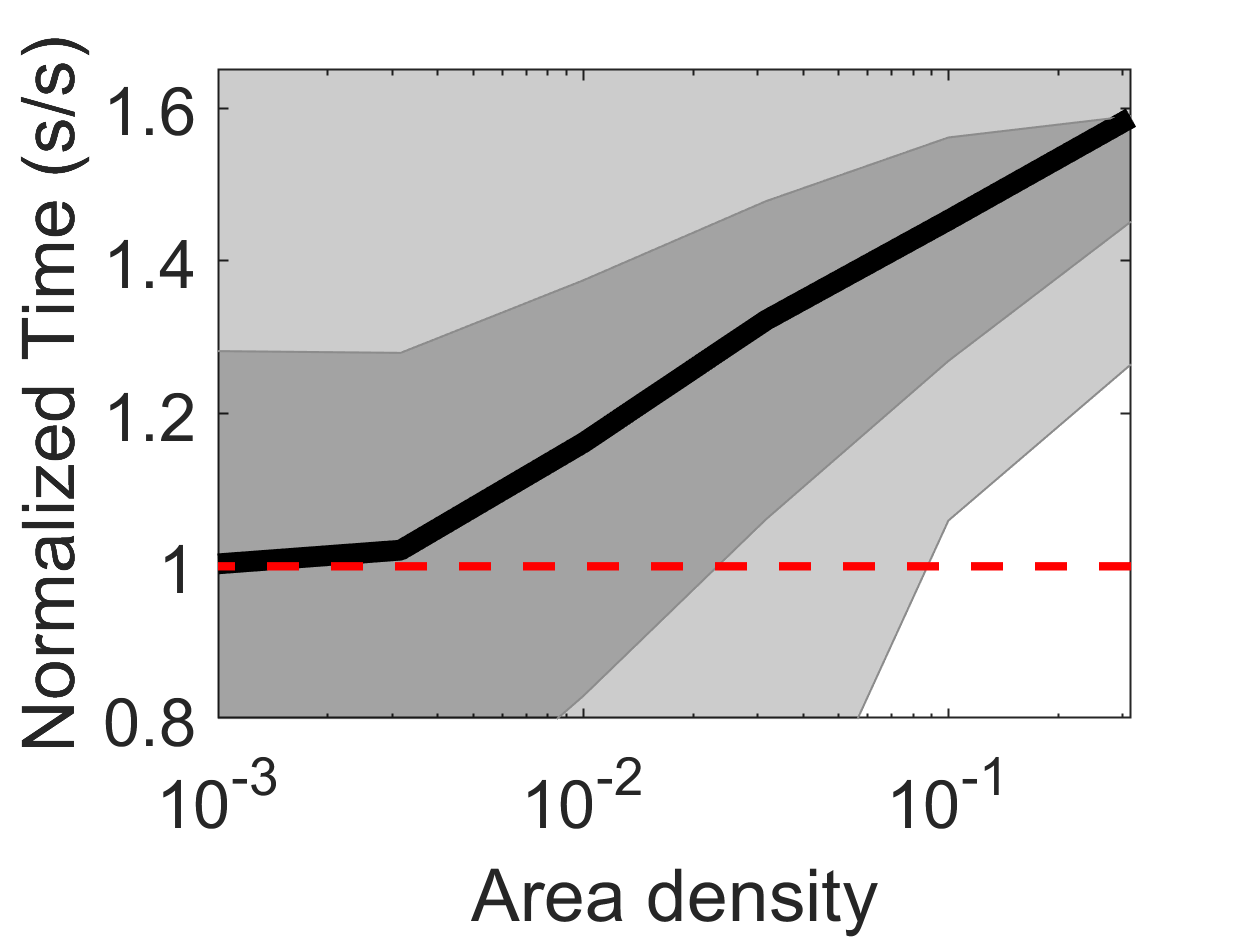}}
  \caption{Time spent in (a) horizontal motion only, (b) vertical motion only, (c) waiting only, and (d) total, using collision resolution via time delay. A y-axis zoomed view of (d) is given in (e). The number of agents was held constant at $n = 100$ and 100 trials were run at each density. Individual data for each agent in each trial are plotted as points. The mean, interquartile range (25th to 75th percentile), and full range (0th to 100th percentile) are plotted as a bold black line, dark shaded region, and light shaded region. The vertical axis is linear scaled and the horizontal axis is log scaled with base 10.}
  \label{fig:monteCarloDelays}  
\end{figure}

\begin{figure}[pos=h] 
\centering
    \subfloat[${\eta = 10^{-1/2} \approx 0.316}$\label{fig:example_N_100_density_0p3162}]{%
      \includegraphics[width=1.6in]{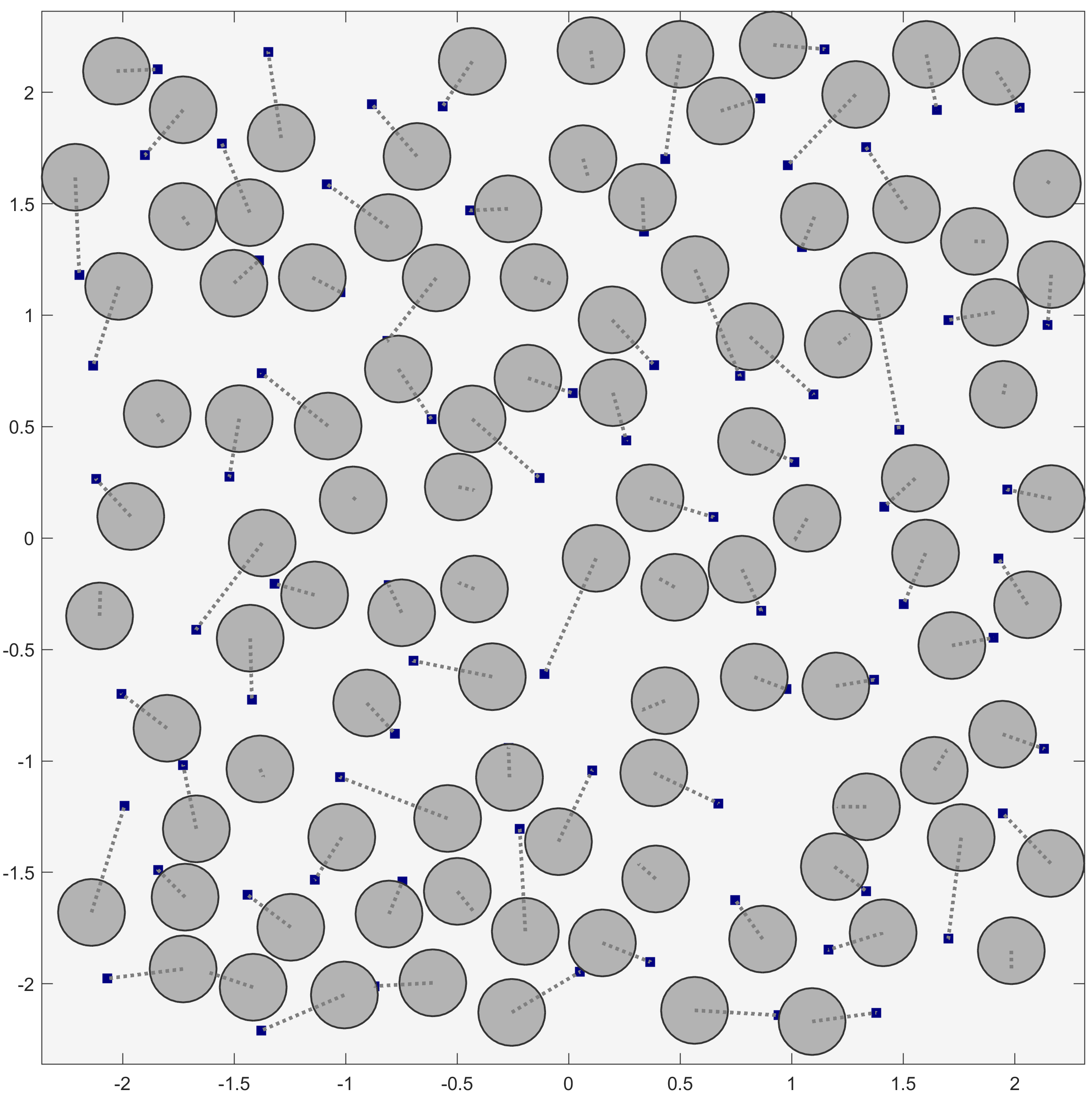}}
    \hfill
    \subfloat[${\eta = 10^{-3/2} \approx 0.0316}$\label{fig:example_N_100_density_0p03162}]{%
      \includegraphics[width=1.6in]{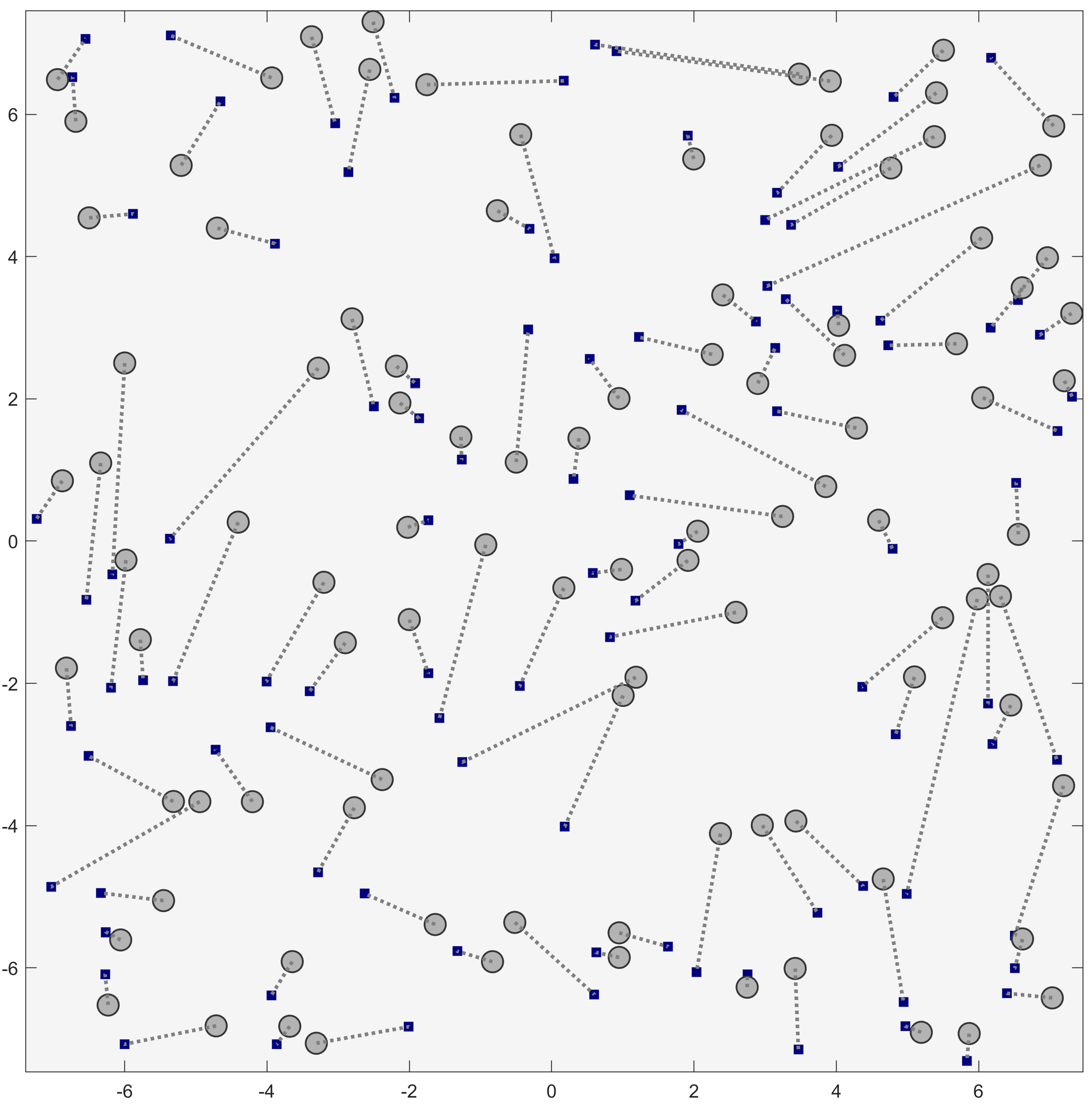}}  
    \caption{Example trajectories using collision resolution via time delays with 100 agents at two different area densities ($\eta$).}
    \label{fig:example_N_100}  
\end{figure}

From Fig. \ref{fig:monteCarloDelays}, for this class of random scenarios, it is evident that as the density increases, the total time taken increases as a larger portion of the time is spent moving vertically and waiting. From the zoomed portion, it is evident that the average induced degradation is manageable, being virtually negligible at low agent densities and peaking at around $60\%$ worse than the lower bound at an agent density of ${\eta = 10^{-1/2} \approx 0.316}$ which represents a highly congested scenario as can be seen in Figure \ref{fig:example_N_100}.

\begin{figure}[pos=h] 
\centering
  \subfloat[Horizontal motion time\label{fig:monteCarloAltitudes_a}]{%
      \includegraphics[width=1.6in]{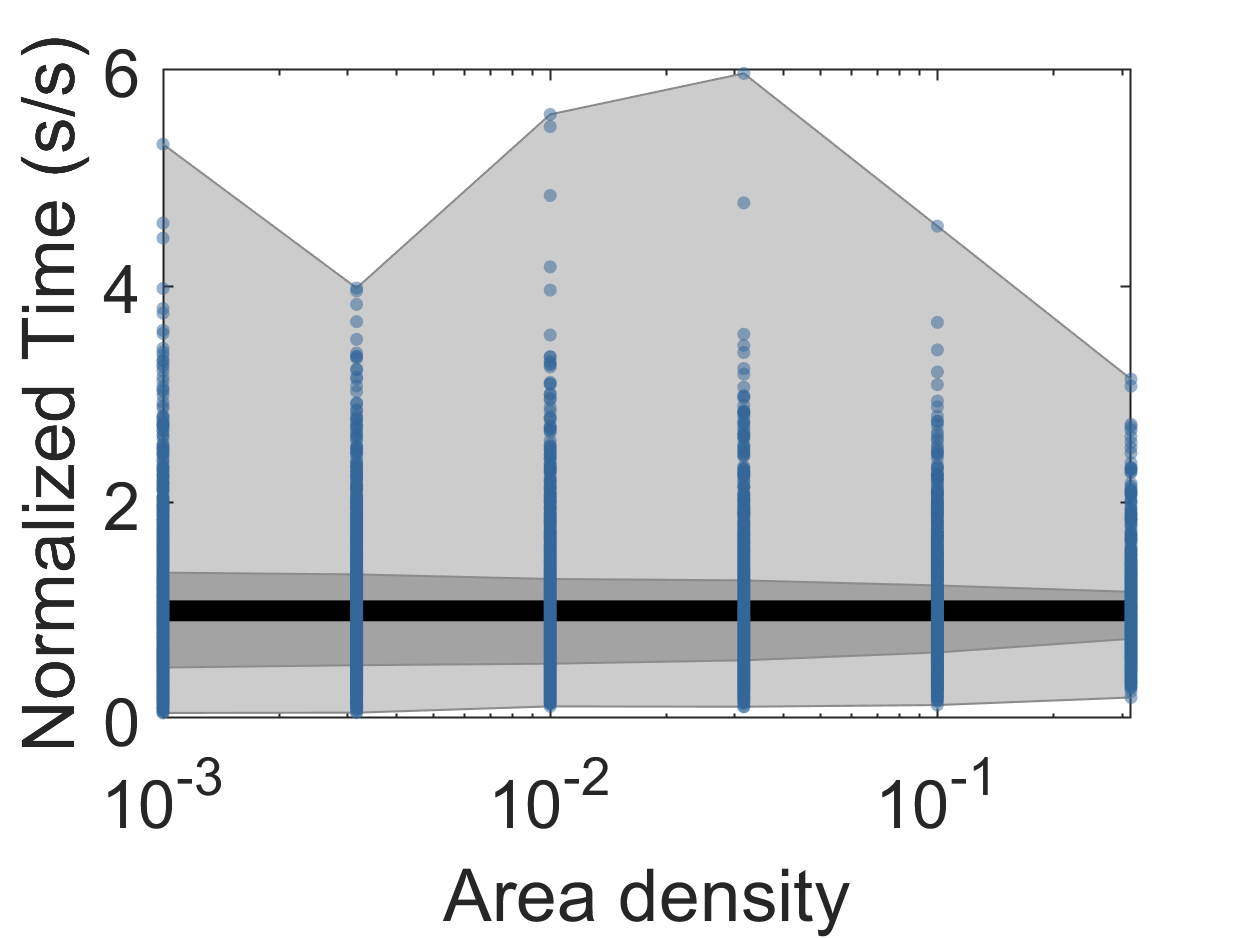}}
    \hfill
  \subfloat[Vertical motion time\label{fig:monteCarloAltitudes_b}]{%
      \includegraphics[width=1.6in]{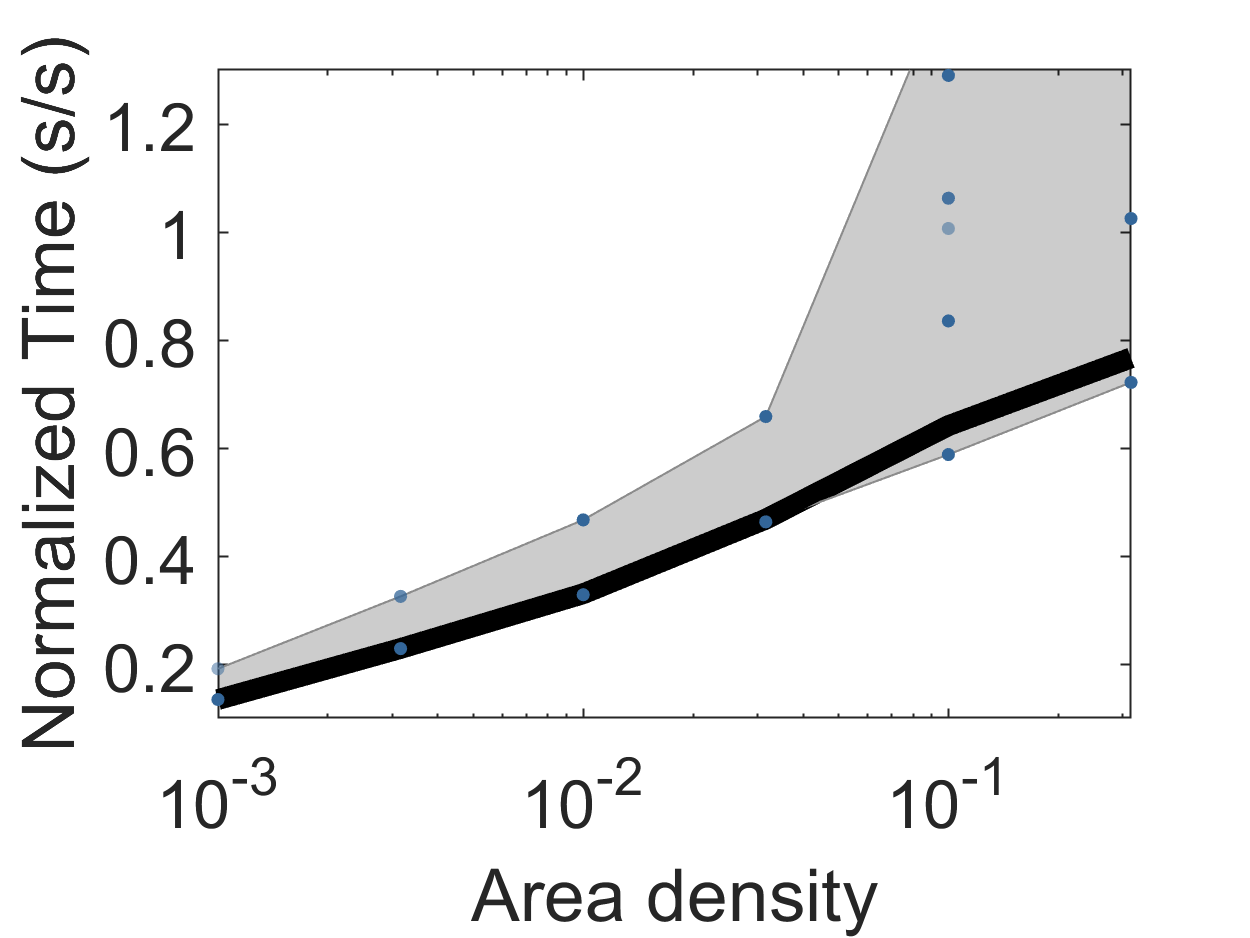}}  
    \hfill
  \subfloat[Waiting time\label{fig:monteCarloAltitudes_c}]{%
      \includegraphics[width=1.6in]{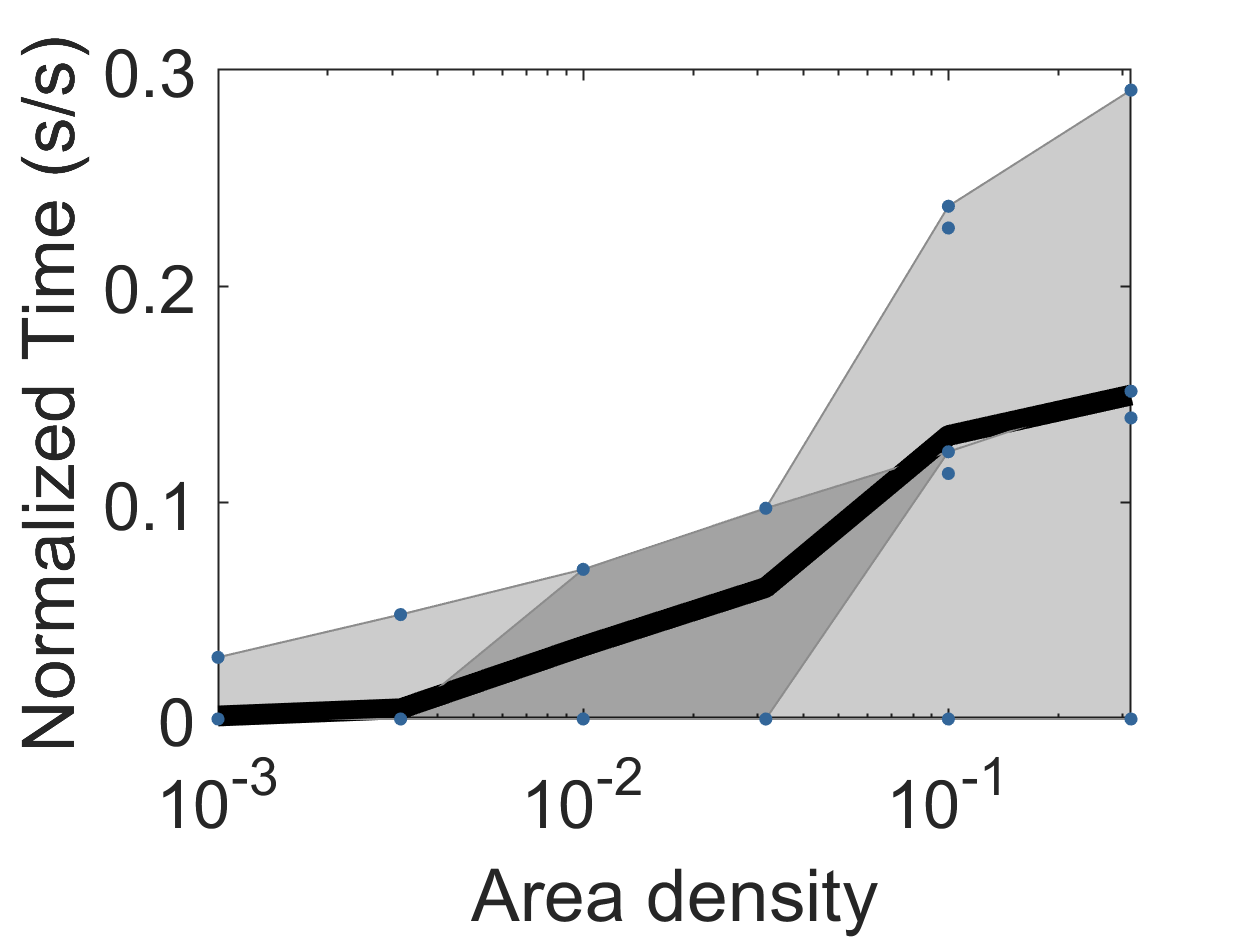}}
    \hfill
  \subfloat[Total time \label{fig:monteCarloAltitudes_d}]{%
      \includegraphics[width=1.6in]{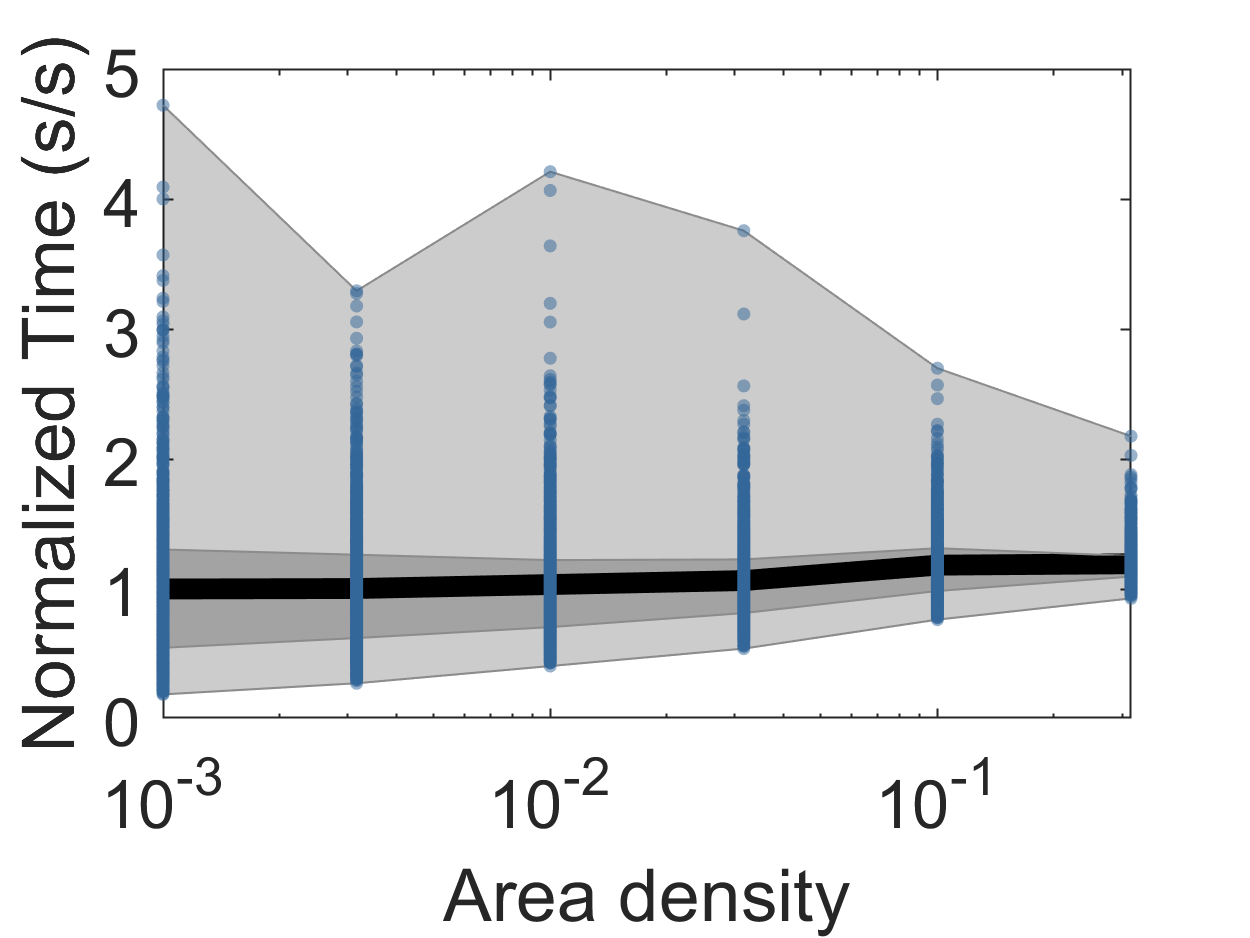}}
    \hfill
  \subfloat[Total time, zoomed \label{fig:monteCarloAltitudes_e}]{%
      \includegraphics[width=1.6in]{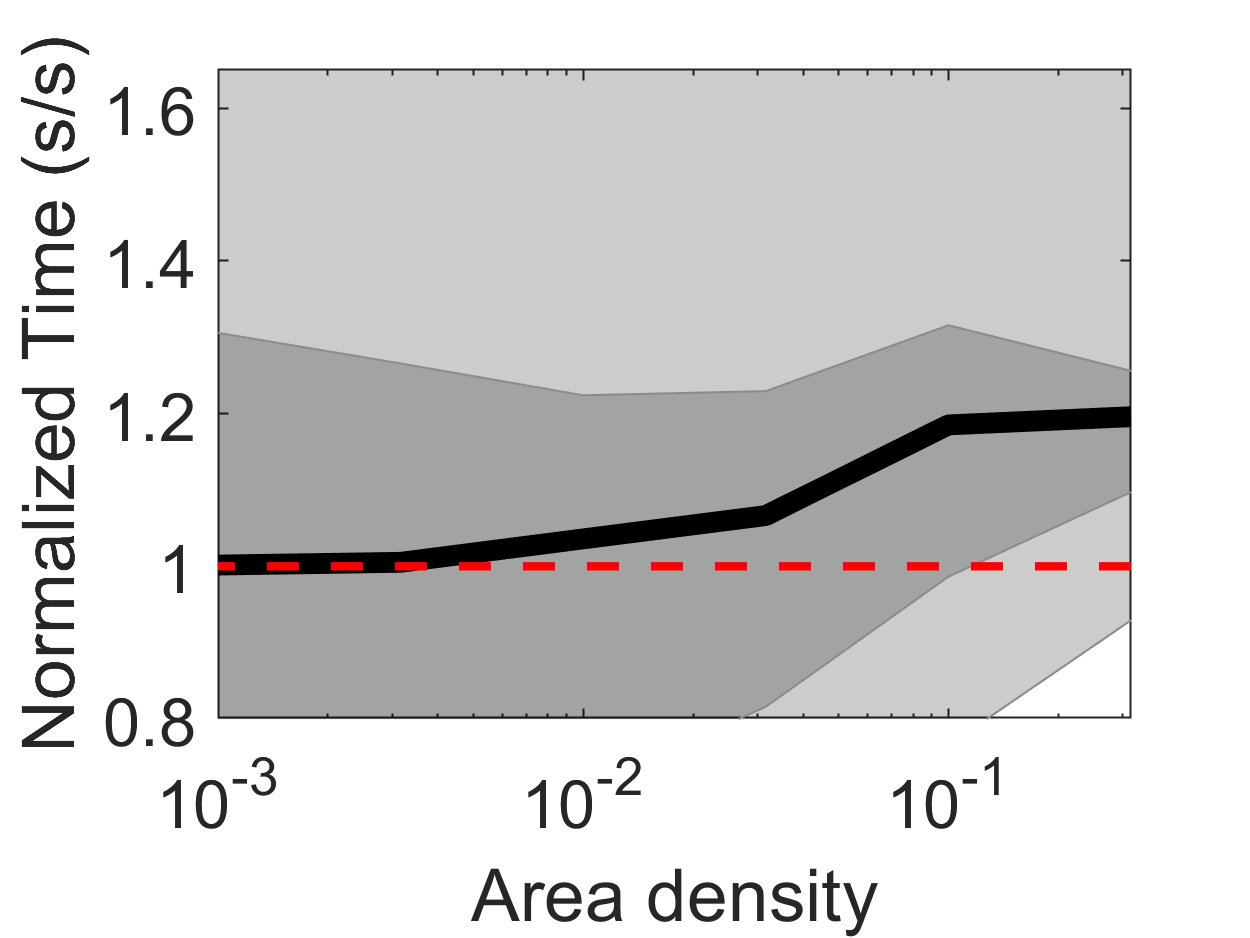}}
  \caption{Time spent in (a) horizontal motion only, (b) vertical motion only, (c) waiting only, and (d) total, using collision resolution via altitude assignment. A y-axis zoomed view of (d) is given in (e). The number of agents was held constant at $n = 100$ and 100 trials were run at each density. Individual data for each agent in each trial are plotted as points. The mean, interquartile range (25th to 75th percentile), and full range (0th to 100th percentile) are plotted as a bold black line, dark shaded region, and light shaded region. The vertical axis is linear scaled and the horizontal axis is log scaled with base 10.}
  \label{fig:monteCarloAltitudes}  
\end{figure}

From Fig. \ref{fig:monteCarloAltitudes} similar trends are observed as in Fig. \ref{fig:monteCarloDelays} using the time delay collision resolution method, but with less time spent waiting, more time spent in vertical motion, and less time spent in total with the average induced degradation again virtually negligible at low agent densities and peaking at around $20\%$ worse than the lower bound at a density of ${\eta = 10^{-1/2} \approx 0.316}$.

\begin{figure}[pos=h] 
\centering
    \subfloat[\label{fig:conl1000_1}]{%
      \includegraphics[width=1.6in]{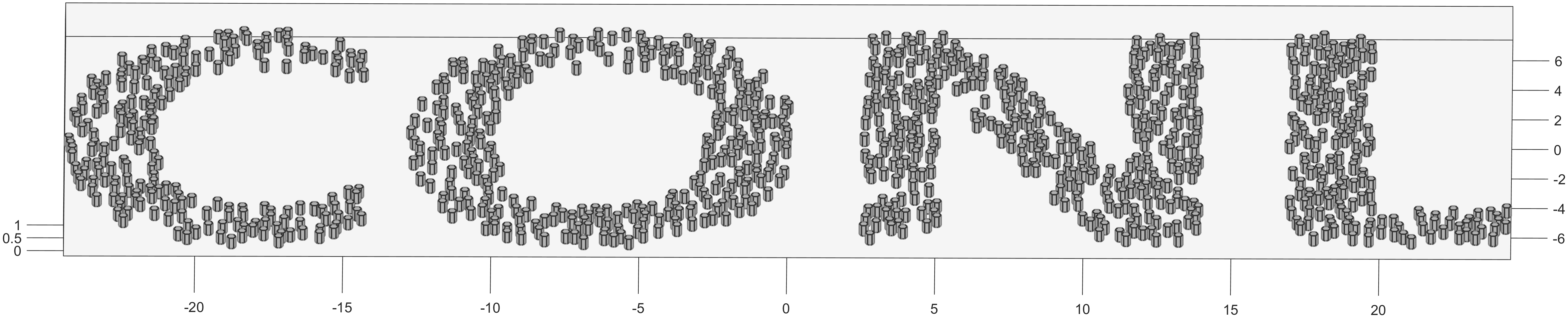}}
    \hfill
    \subfloat[\label{fig:conl1000_2}]{%
      \includegraphics[width=1.6in]{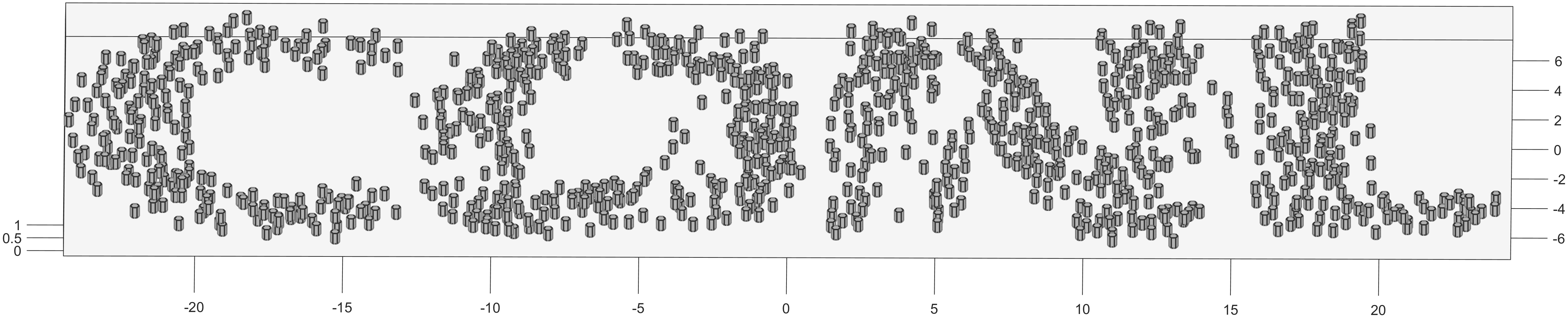}}  
    \hfill
    \subfloat[\label{fig:conl1000_3}]{%
      \includegraphics[width=1.6in]{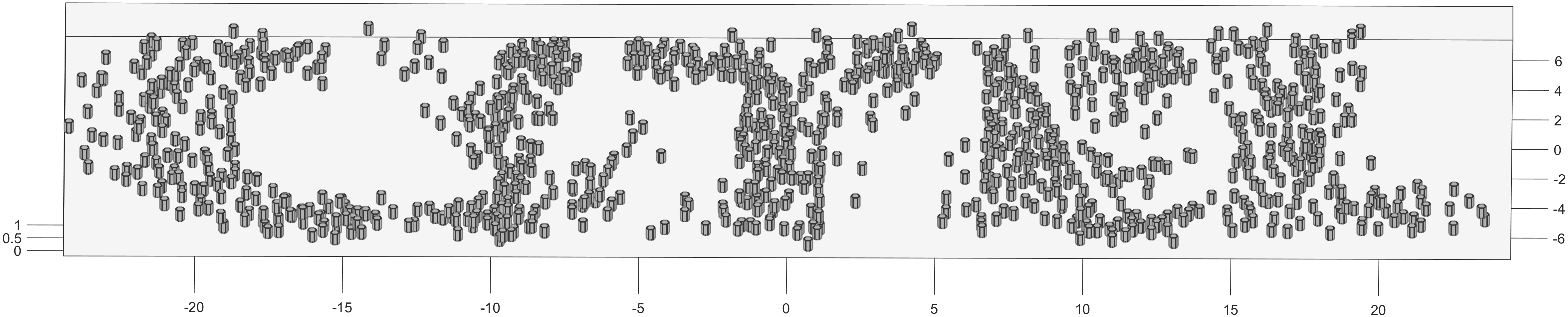}}
    \hfill
    \subfloat[\label{fig:conl1000_4}]{%
      \includegraphics[width=1.6in]{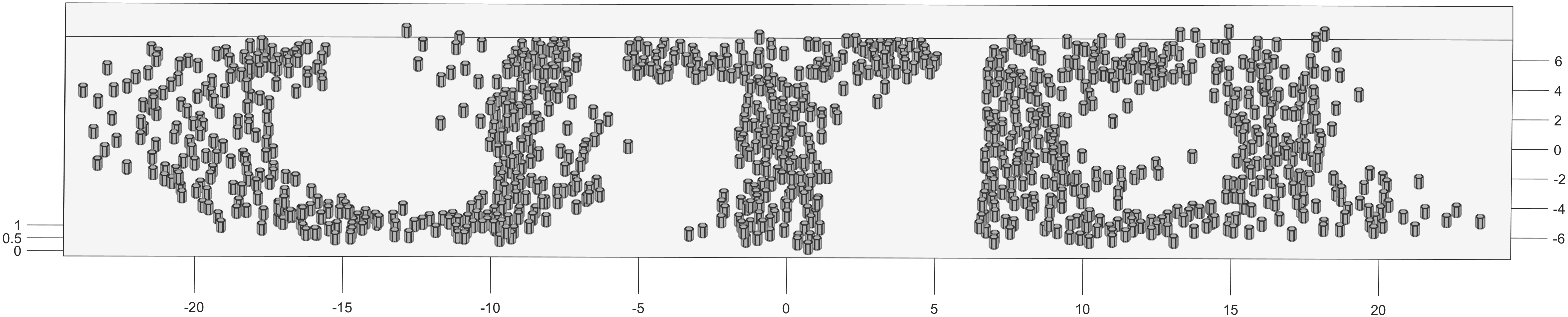}}
    \hfill
    \subfloat[\label{fig:conl1000_5}]{%
      \includegraphics[width=1.6in]{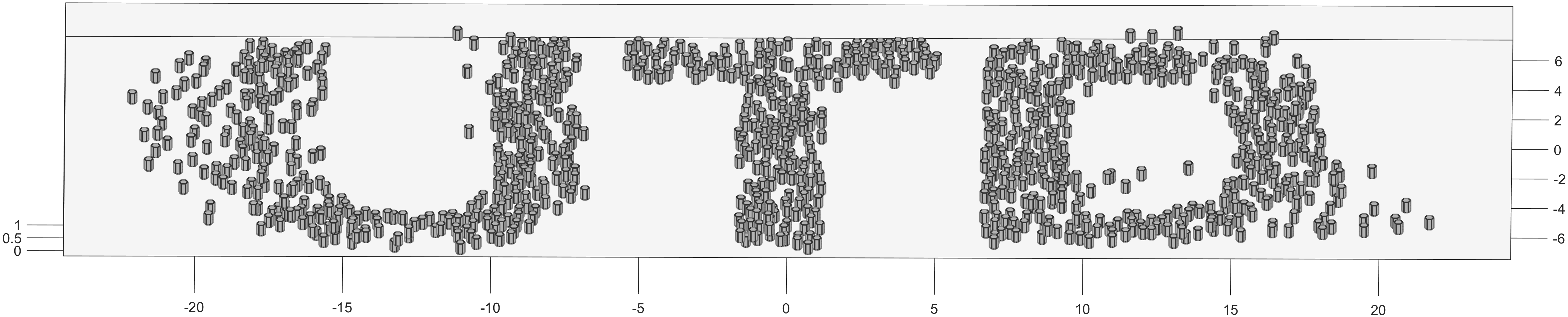}}
    \hfill
    \subfloat[\label{fig:conl1000_6}]{%
      \includegraphics[width=1.6in]{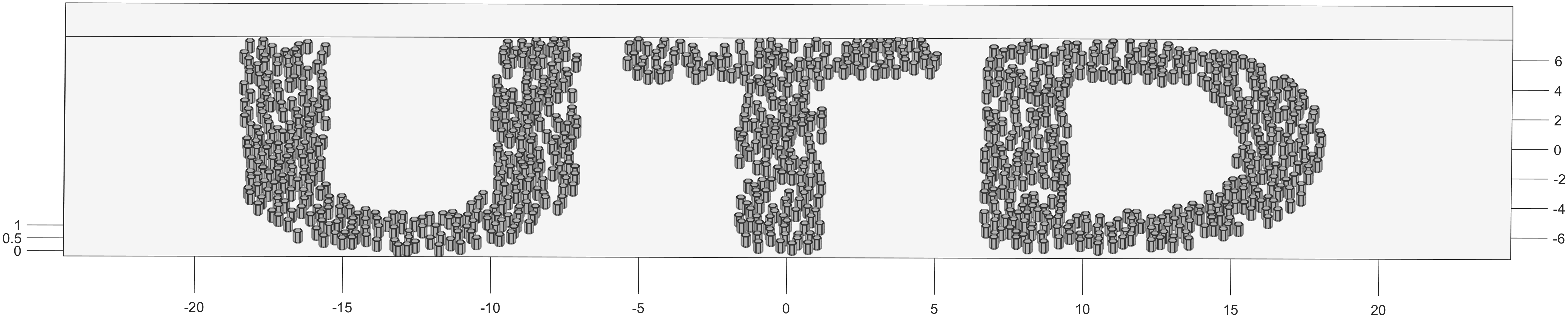}}
    \caption{Example of 1000 agents spelling out the letters of the authors' lab and university by following collision-free trajectories generated by the proposed algorithm using collision resolution via time delays.}
    \label{fig:example_N_1000}  
\end{figure}

In Figures \ref{fig:example_N_100} and \ref{fig:example_N_1000} example trajectories generated by the proposed algorithm are shown. Figure \ref{fig:example_N_100} gives a visualization of the scale of the area density, while Figure \ref{fig:example_N_1000} demonstrates the ability of the proposed algorithm to plan trajectories for a large number of vehicles navigating between arbitrary locations.

\subsubsection{Performance relative to alternate methods}

For the purpose of comparing the proposed method to alternate methods e.g. that of \cite{turpin2014}, define the characteristic time $t_c$ which is the time an agent would take to traverse the longest horizontal straight-line path within the space, which for a square space has length $\sqrt{2}S$. For a given trajectory plan, denote the time spent by agent $i$ in horizontal motion and in waiting respectively as $t_{h,i}$ and $t_{w,i}$.
Also define the characteristic normalized time in horizontal motion and in waiting $t_p$ as 
\begin{equation*}
\widetilde{t}_p = \frac{\frac{1}{n} \sum_{i=1}^n (t_{h,i}+t_{w,i})}{t_c} .
\end{equation*}

For simplicity, collisions were allowed for simulations using the approach of \cite{turpin2014} to avoid imposing the $2\sqrt{2}R$ separation condition required for that approach to possess collision-free guarantees; this was conservative in the sense that the relative performance of the proposed methods relative to \cite{turpin2014} was only degraded by this assumption. Also, for simplicity simple 1-degree (constant speed) polynomials were used for trajectory generation.

\begin{figure}[pos=h] 
\centering
    \includegraphics[width=3in]{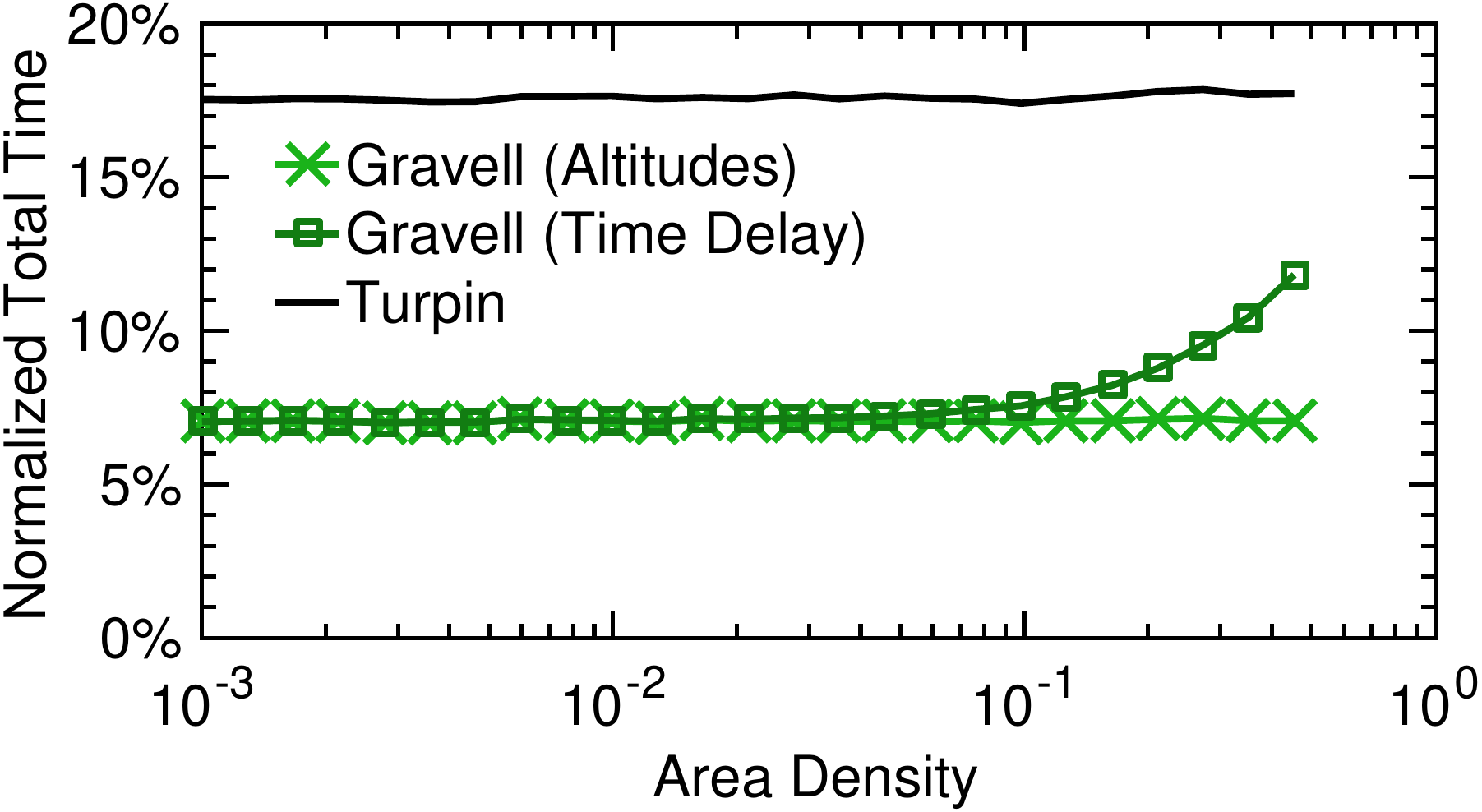}
    \caption{Normalized average flight times using the proposed collision resolution methods vs the approach of \cite{turpin2014}. The mean value from 1000 trials over all $n=100$ agents is plotted.}
    \label{fig:TTmonteCarlo}  
\end{figure}

With respect to the $t_p$ metric, plotted in Fig.~\ref{fig:TTmonteCarlo}, the proposed altitudes approach gave the best results for all densities. At low densities, the proposed time delay approach gave nearly the same performance as the altitudes approach as a consequence of small time delays which vanish as the density goes to zero. At higher agent densities, the time delay approach result began to increase as the physical extent of the agents became more influential. Both proposed approaches performed better at all densities than the approach of \cite{turpin2014}.

\subsection{Computation Time} \label{sec:compSpeed} 
\begin{figure}[pos=h]
    \centering     
    \subfloat[High density ${\eta = 10^{-1/2} \approx 0.316}$\label{fig:figure_monteCarloDelays_compute_time_high_density}]{%
      \includegraphics[width=3in]{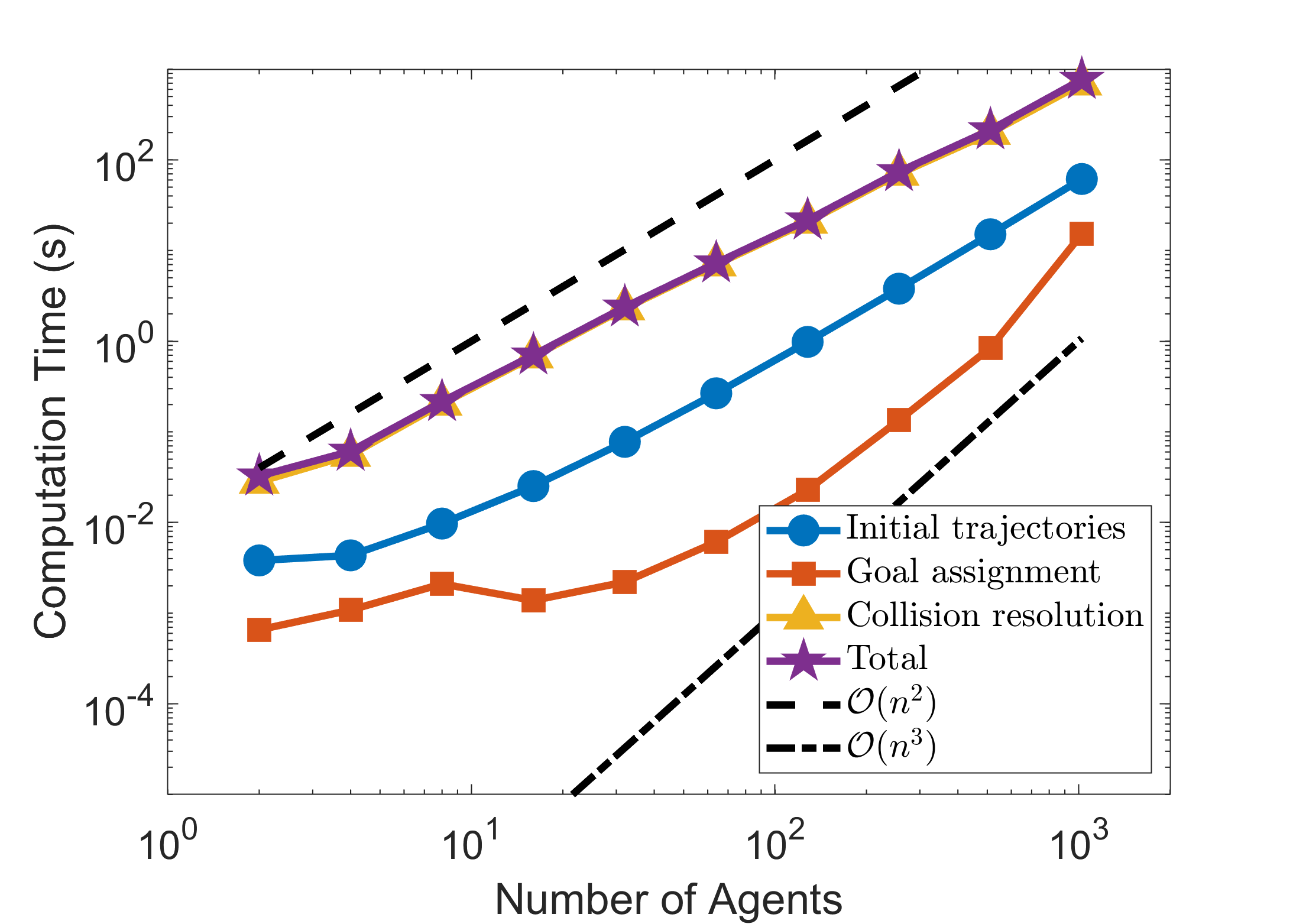}}
    \hfill
    \subfloat[Low density ${\eta = 10^{-3/2} \approx 0.0316}$\label{fig:figure_monteCarloDelays_compute_time_low_density}]{%
      \includegraphics[width=3in]{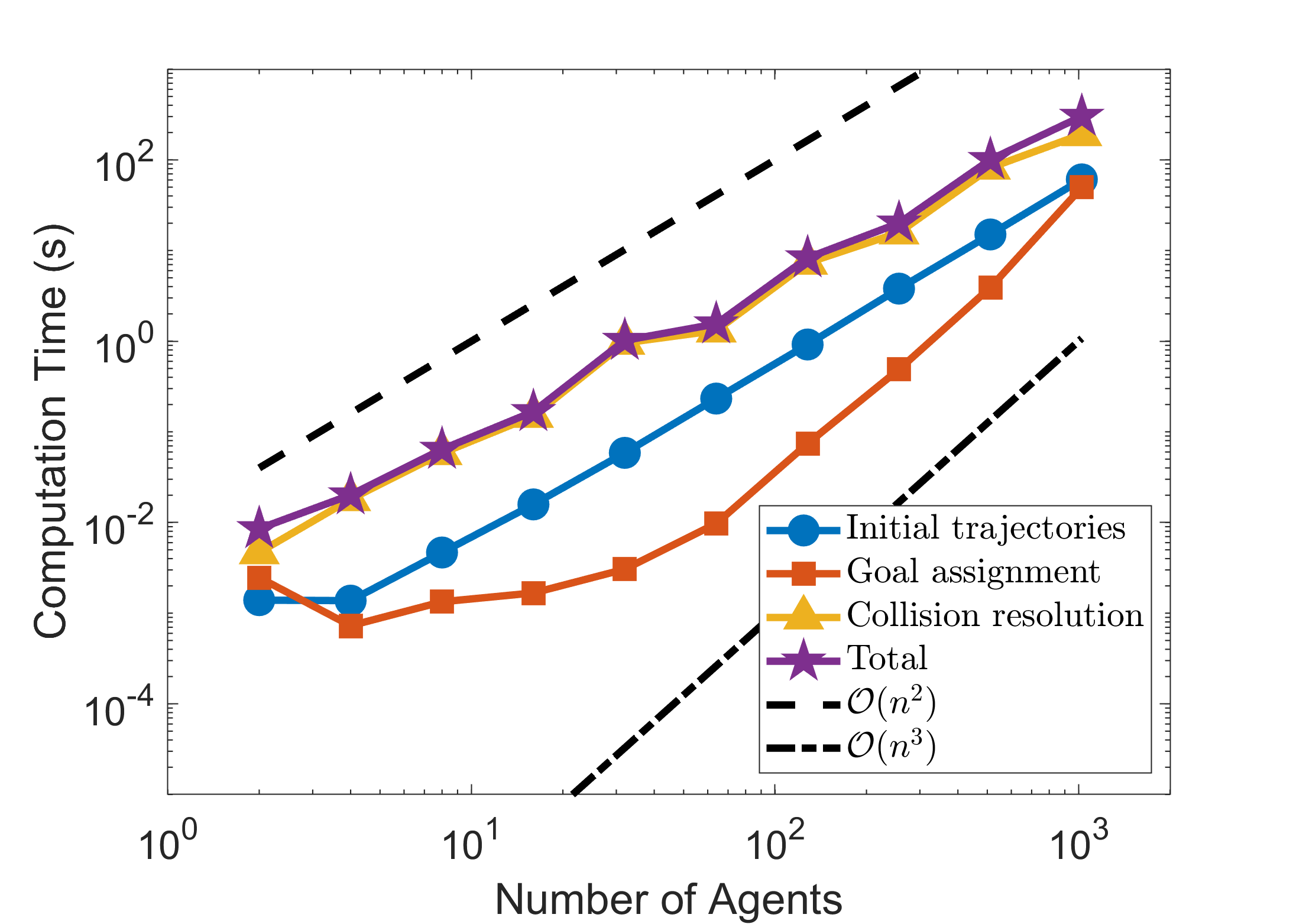}}  
    \caption{Computational time as a function of number of agents using various trajectories using the time delay collision resolution method. The area density was held constant at (a) ${\eta = 10^{-1/2} \approx 0.316}$ and (b) ${\eta = 10^{-3/2} \approx 0.0316}$. The number of agents varied from 2 to 1024 at each power of 2. The mean value from 10 trials at each number of agents is shown.}
    \label{fig:CompTime_time_delay}      
\end{figure}

\begin{figure}[pos=h]
    \centering     
    \subfloat[${\eta = 10^{-1/2} \approx 0.316}$\label{fig:figure_monteCarloAltitudes_compute_time_high_density}]{%
      \includegraphics[width=3in]{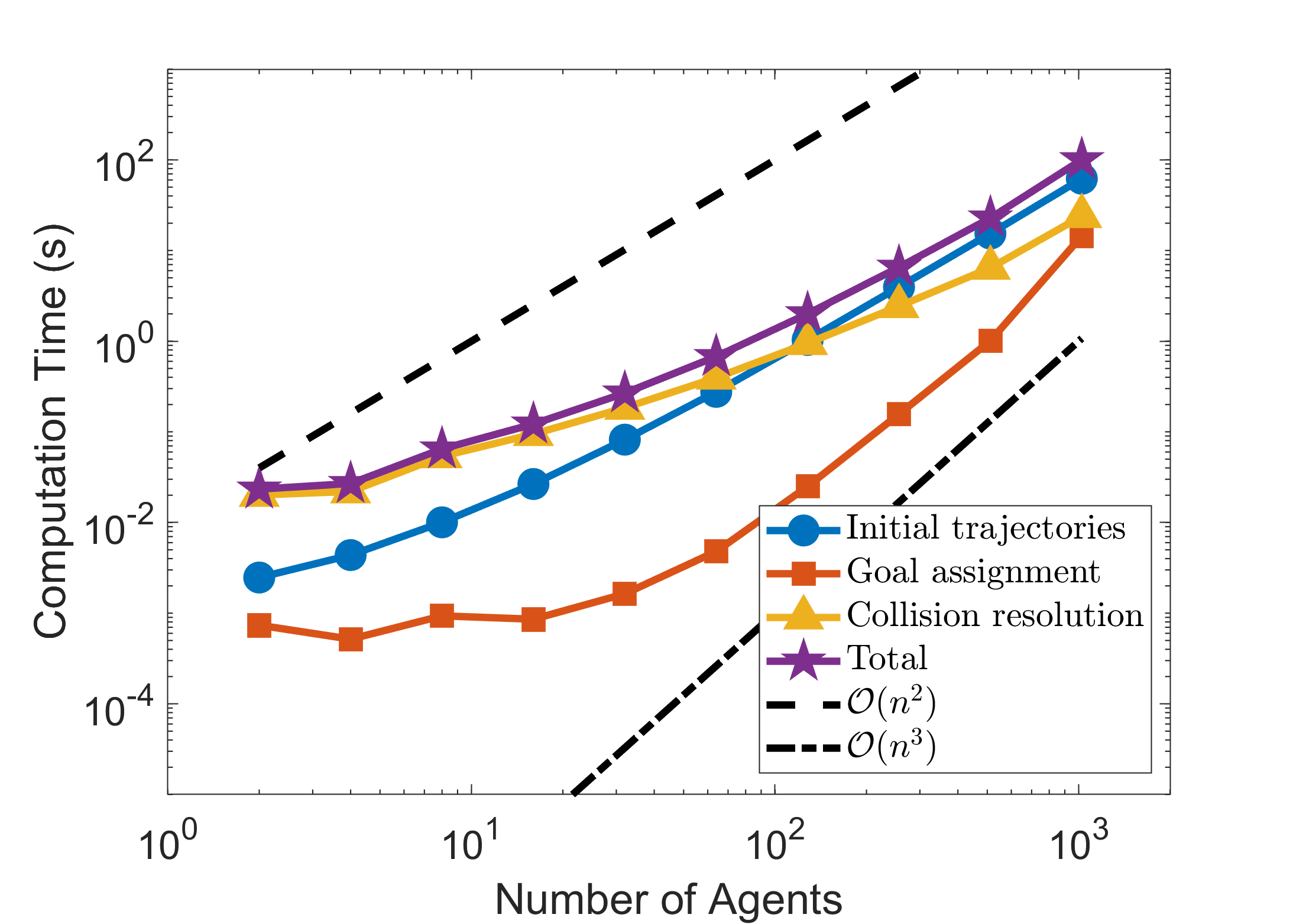}}  
    \caption{Computational time as a function of number of agents using various trajectories using the altitudes collision resolution method. The area density was held constant at ${\eta = 10^{-1/2} \approx 0.316}$. The number of agents varied from 2 to 1024 at each power of 2. The mean value from 10 trials at each number of agents is shown.}
    \label{fig:CompTime_altitudes}      
\end{figure}

Achieving low computation time is an important practical consideration for successful deployment of large robot teams. The simulations were implemented in MATLAB running on a desktop with an AMD Ryzen 7 2700X eight-core processor running at 3.7GHz. The results are given in Figures \ref{fig:CompTime_time_delay} and \ref{fig:CompTime_altitudes}, where reasonable computation times for large teams are observed. Explicitly optimizing the code for performance or parallelization could decrease the computation times even further. The overall computation is split into three major segments: the generation of initial trajectories which occurs when finding the cost matrix for input into the goal assignment, the Hungarian algorithm which actually does the goal assignment, and the combined collision detection and resolution steps. This encompasses nearly all of the computations, with the exception of the base polynomial generation and some post-processing steps which together take negligible time to execute.

The goal assignment computation time grew as $\mathcal{O}(n^3)$ as expected from a standard computational complexity analysis \cite{munkres1957}. The trajectory generation and collision resolution steps grew only as $\mathcal{O}(n^2)$ since the average number of pairwise trajectories and collisions grew with the number of pairs of agents. At ever higher agent numbers it is inevitable that the goal assignment will began to dominate.

\section{Experimental results}
\label{sec:experimental_results}

A series of experiments on physical hardware was performed to validate the performance and safety of the proposed approach. 

\subsection{System description}

A proprietary branch of the Crazyswarm system was used, which encompasses both hardware and software \cite{Preiss2017}. State estimation was accomplished by taking position measurements of infrared (IR) markers on each quadrotor with an external Vicon camera system. The point cloud of these individual position measurements were then resolved into body coordinate frame (state) estimates using the object tracker portion of the Crazyswarm software. These state estimates were then used by the Crazyswarm software to generate feedback control signals, which were then broadcast over wireless radios to the flying vehicles and electrically converted to motor voltages, completing the feedback loop. { The controller used was the standard ``Mellinger'' controller implemented by the Crazyswarm package, which is a modified version of the nonlinear reference-tracking controller proposed by \cite{mellinger2011} which takes advantage of differential flatness of the quadrotor.}
See the documentation at \url{https://github.com/TSummersLab/crazyswarm} for further details of the software and hardware setup.

\begin{figure}[pos=h]
    \centering     
    \includegraphics[width=1.6in]{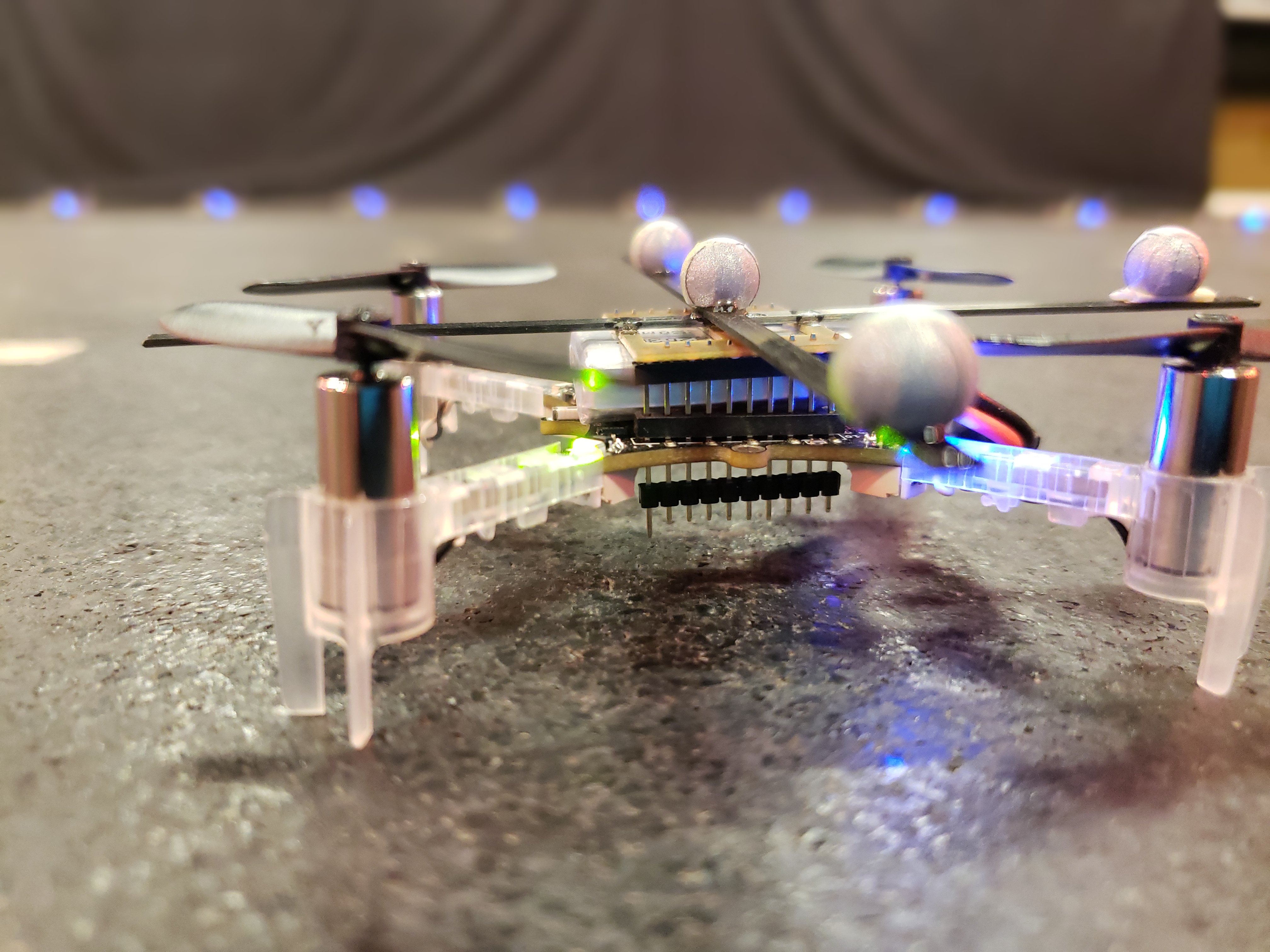}
    \caption{Photograph of a single Crazyflie quadrotor vehicle.}
    \label{fig:crazyflie}  
\end{figure}

\subsection{Trajectory tracking errors}

One immediate practical issue was reference trajectory tracking in the presence of noise; although the planned trajectories could be followed perfectly in the absence of disturbances, the presence of disturbances precludes this possibility. Natural sources of disturbances included ambient air currents from air conditioner vents, downwash from other agents, ground aerodynamic effects, other unmodeled dynamics, and sensor (camera) noise. By simple enlargement of the collision volumes and ensuring bounds on the position error of all agents, collision avoidance remained guaranteed. Feedback control within the Crazyswarm package ensured the position deviation of each agent from the desired position remained small at all times. In particular, it was found from the experiments that during all trajectory traversals that the radial position error was bounded by 8 cm and the vertical position error was bounded by 7cm; the plot in \ref{fig:position_error} demonstrates satisfaction of these bounds. Additionally, it was found that the effects of downwash were accounted for by extending the bottom of the collision cylinder by an additional 20cm. Thus, by choosing a collision cylinder with dimensions enlarged by these amounts relative to the physical dimensions of the quadrotor i.e. diameter of $14 + 2 \times 8 = 30$ cm and height of $4 + 2 \times 7 + 20 = 40$ cm, the vehicle was guaranteed to always be strictly contained within the collision volume, maintaining collision avoidance guarantees. This can be observed from from Figure \ref{fig:compare_sim_exp_X20_clearance}; the trajectory generation tightly respected the collision constraints, as the minimum clearance approached zero without becoming negative. Likewise, during the physical experiment the agents did not experience any collisions, as evidenced by the strictly positive clearance. This was true for all experiments.

\subsection{Experiment description and findings}
\begin{figure}[pos=h] 
\centering
\subfloat[Start\label{fig:X20_top_landed_start}]{%
      \includegraphics[width=1.6in]{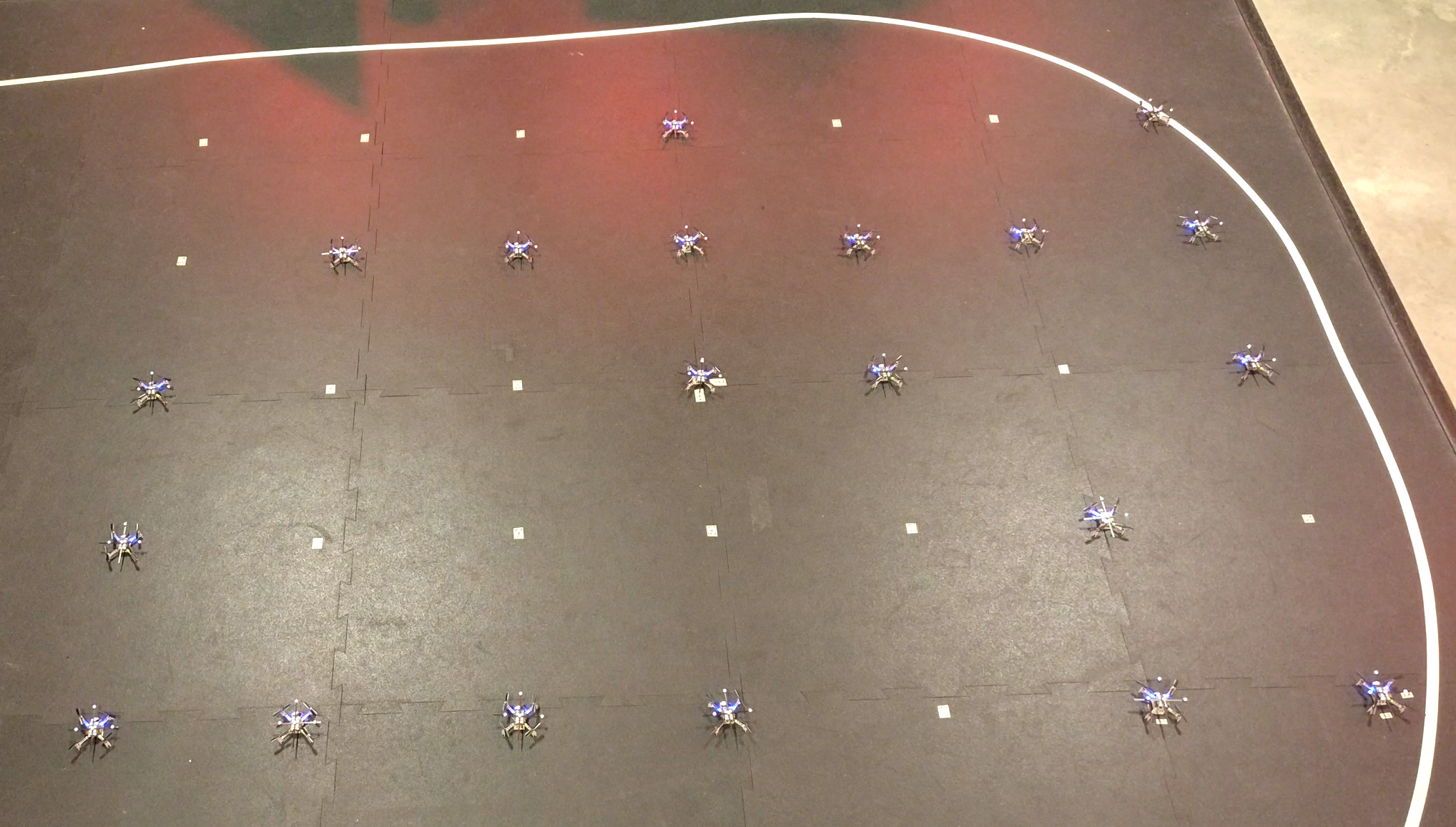}}
    \hfill
      \subfloat[End\label{fig:X20_top_landed_end}]{%
      \includegraphics[width=1.6in]{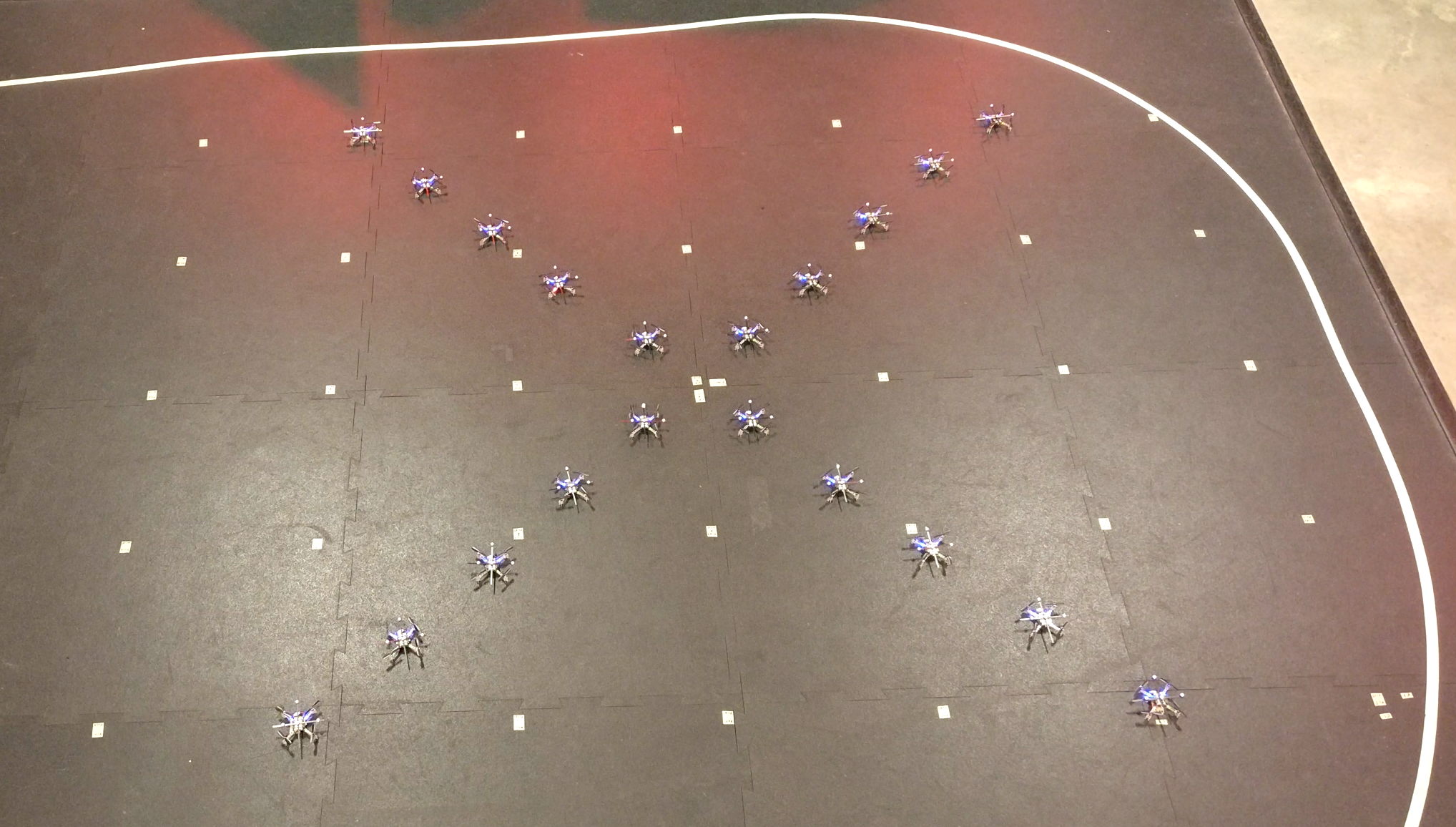}}  
    \hfill       
  \subfloat[Mid-flight\label{fig:X20_side_midflight}]{%
      \includegraphics[width=2.4in]{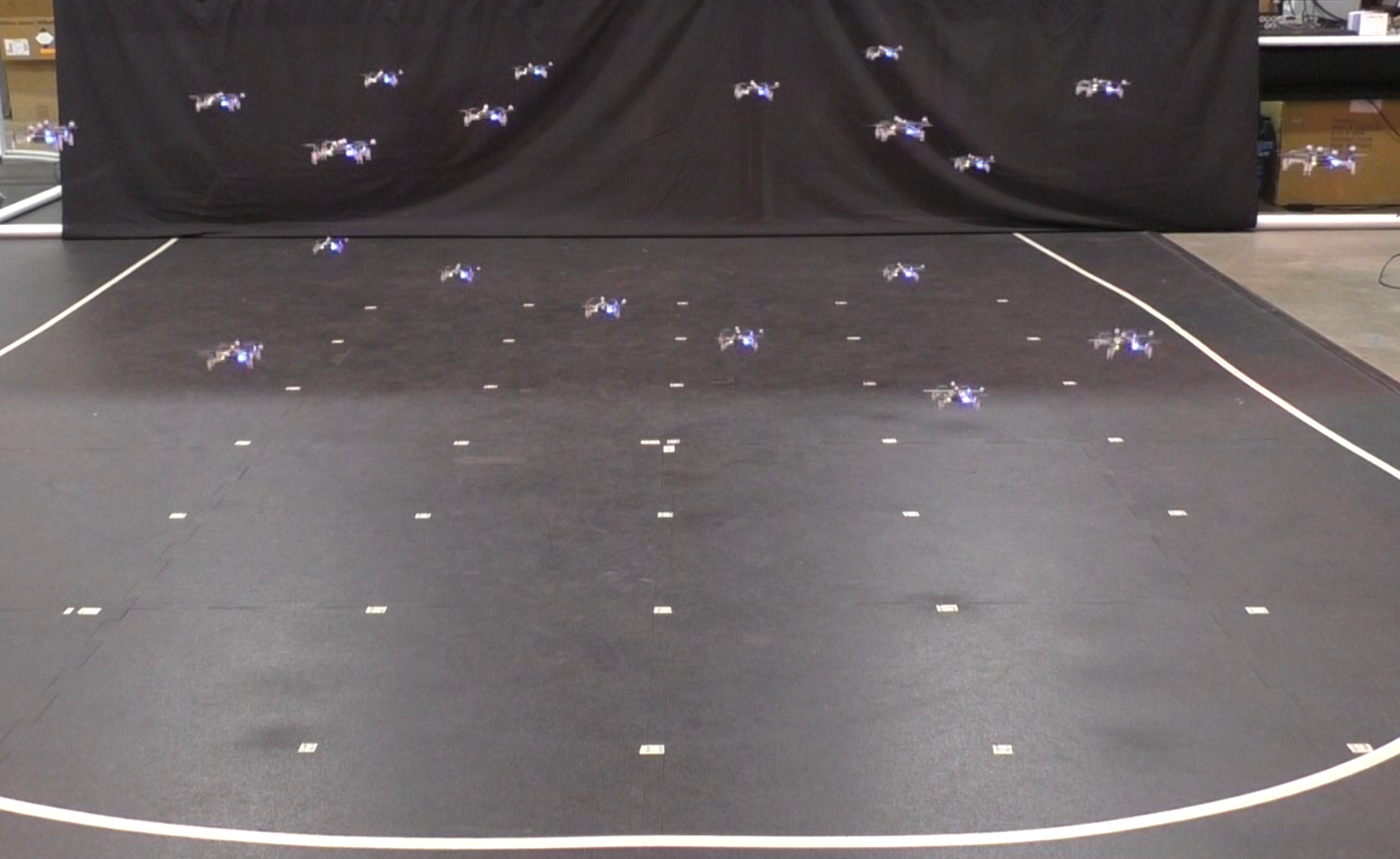}}    
  \caption{Photographs of experimental setup at the (a) start configuration, top view, (b) end configuration, top view, and (c) mid-flight, side view.} 
  \label{fig:experiment_photo}  
\end{figure}

\begin{figure}[pos=h]
    \centering     
    \includegraphics[width=3in]{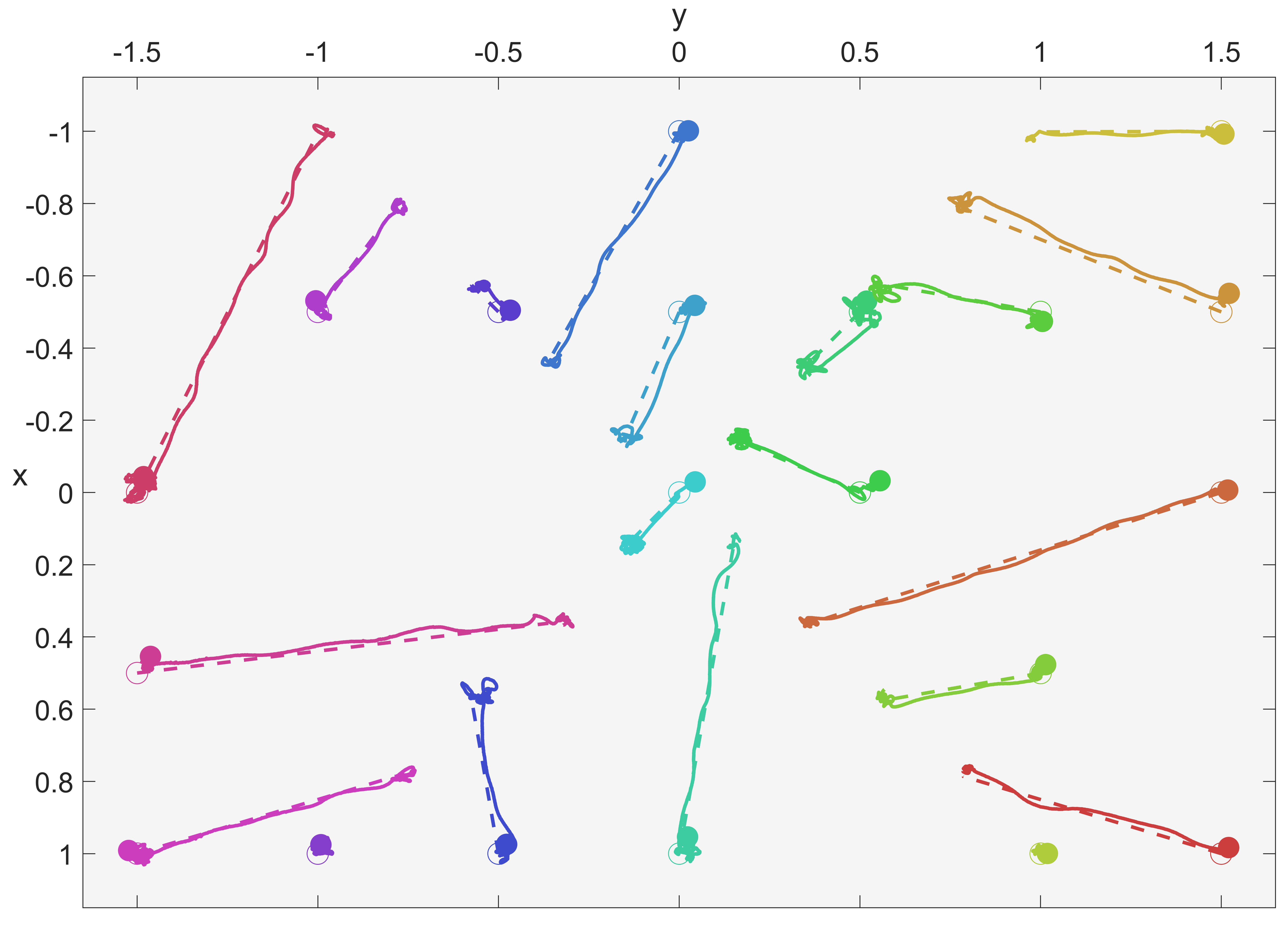}
    \caption{Top-down view of agent centers (filled dots) and trajectories at the beginning of the X20 experiment. Desired trajectory paths shown as dashed lines and actual paths realized by the physical vehicles are shown as solid lines.}
    \label{fig:compare_sim_exp_O24_topview}  
\end{figure}

One experiment (``X20'') is now presented with $n = 20$ agents moving in a 2m by 3m room from start locations randomly selected from a grid with 0.5m spacing to goals arranged in an ``X'' configuration roughly 2.8m across the widest section; see Figures \ref{fig:experiment_photo} and \ref{fig:compare_sim_exp_O24_topview}.
In this experiment the time delay collision resolution method was used. This was for the practical reason that the height of the room limited the number of usable altitudes; in outdoor environments the height of the flyable space would be much greater.
The kinematic constraints imposed during trajectory generation were the same as for the simulation results i.e. those listed in Table \ref{tab:table_kinematic_constraints}.

\begin{figure}[pos=h] 
\centering
\subfloat[Simulation\label{fig:compare_sim_exp_X20_sim_clearance}]{%
      \includegraphics[width=1.6in]{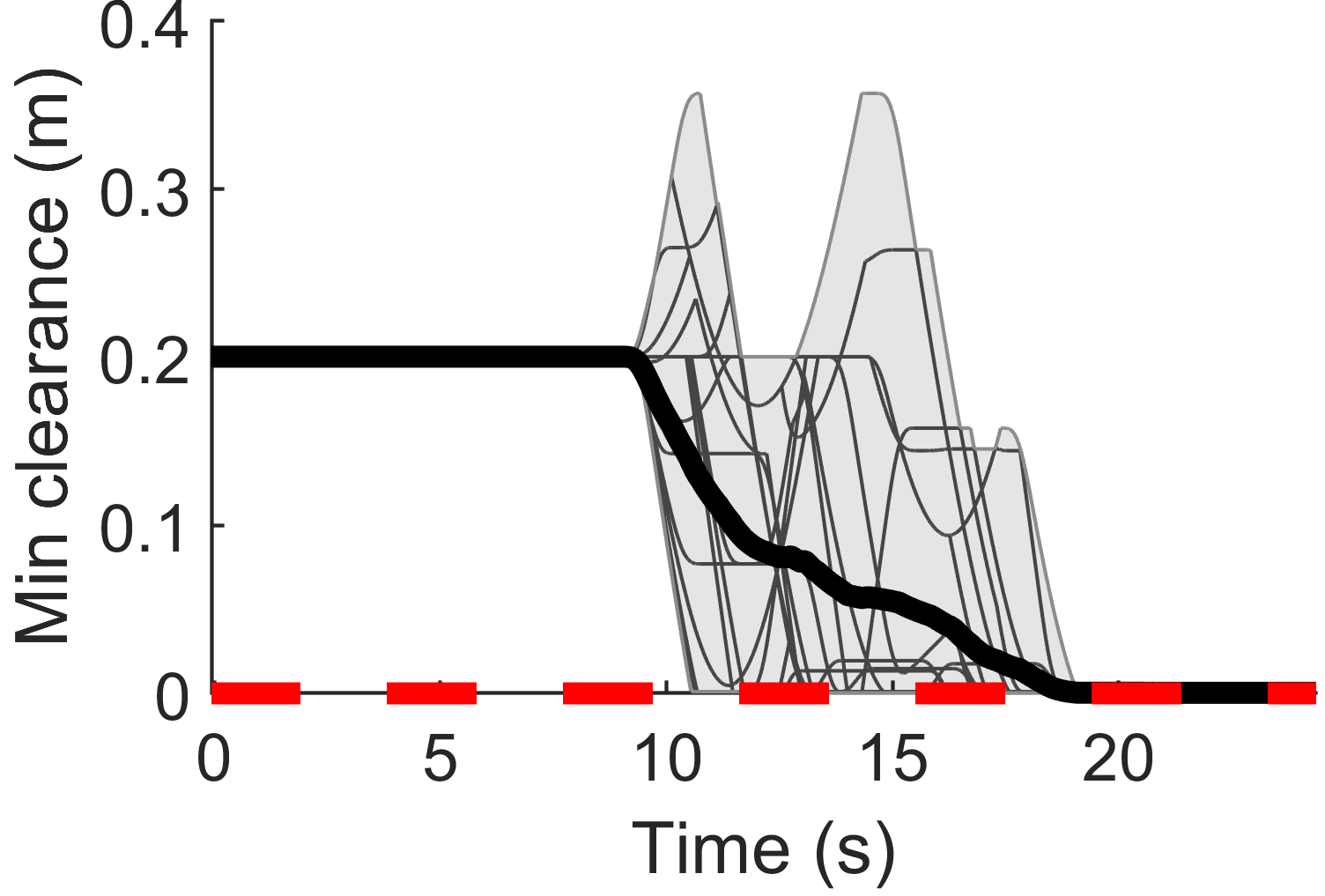}}
    \hfill
  \subfloat[Experiment\label{fig:compare_sim_exp_X20_exp_clearance}]{%
      \includegraphics[width=1.6in]{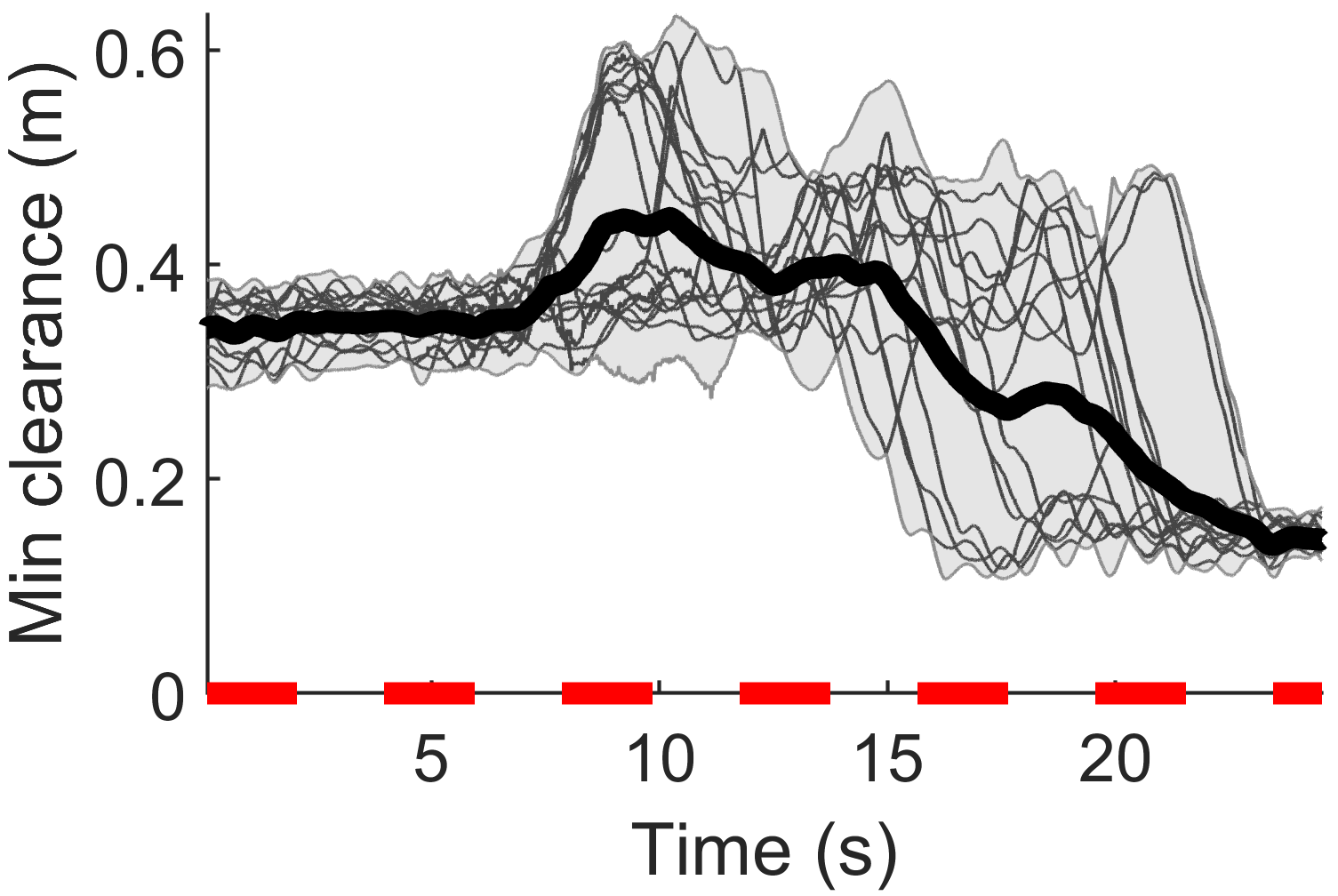}}  
  \caption{Minimum clearance between each agent and all other agents vs time, showing (a) clearance using the enlarged collision volumes on the disturbance-free simulation trajectories (b) clearance using the actual vehicle boundary volume on the noisy realized physical paths. The minimum clearance of each agent is plotted as a thin line, the mean as a thick line, the range between minimum and maximum shaded in grey, and zero as a dashed line.} 
  \label{fig:compare_sim_exp_X20_clearance}  
\end{figure}

\begin{figure}[pos=h] 
\centering
\subfloat[Horizontal\label{fig:position_error_horz}]{%
      \includegraphics[width=1.6in]{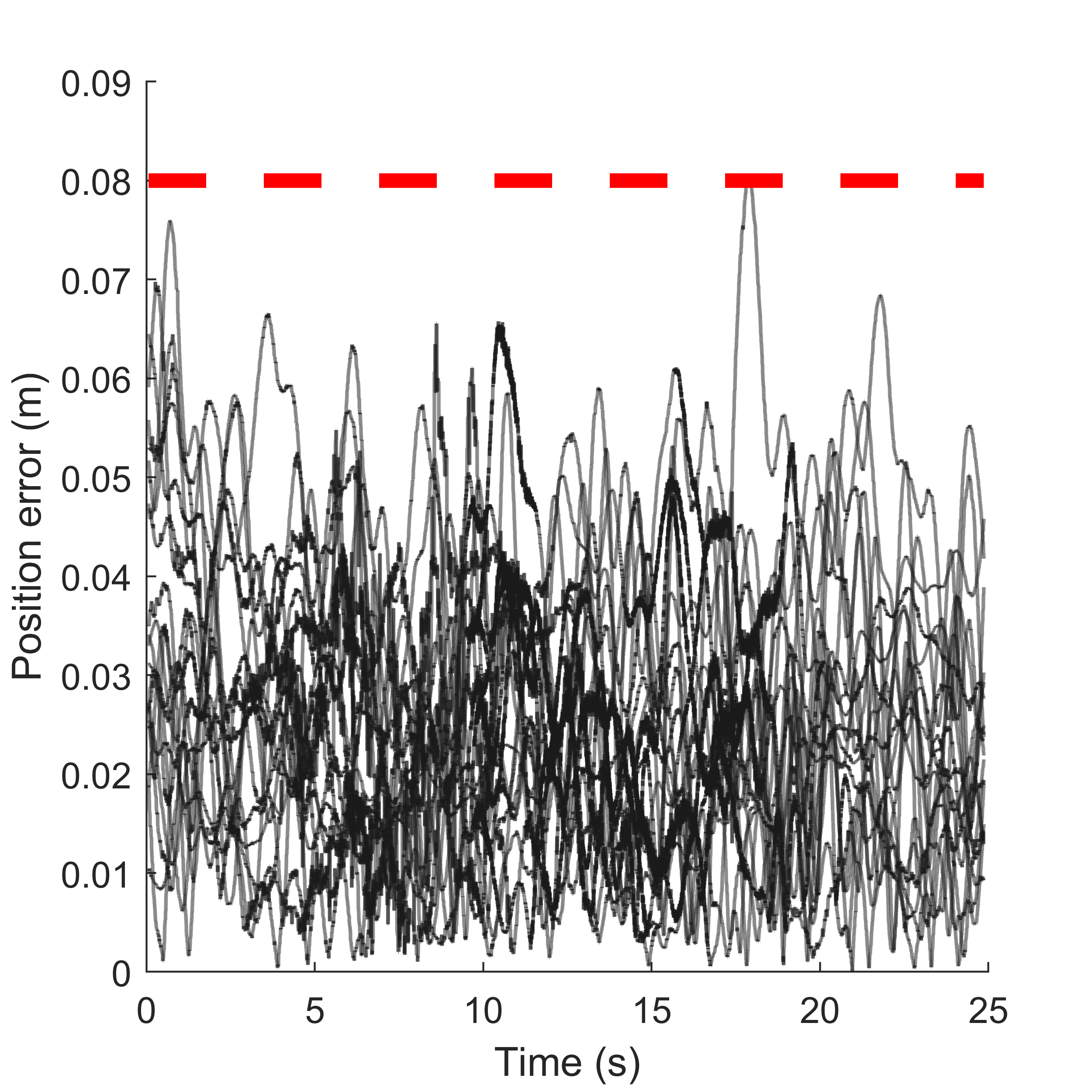}}
    \hfill
  \subfloat[Vertical\label{fig:position_error_vert}]{%
      \includegraphics[width=1.6in]{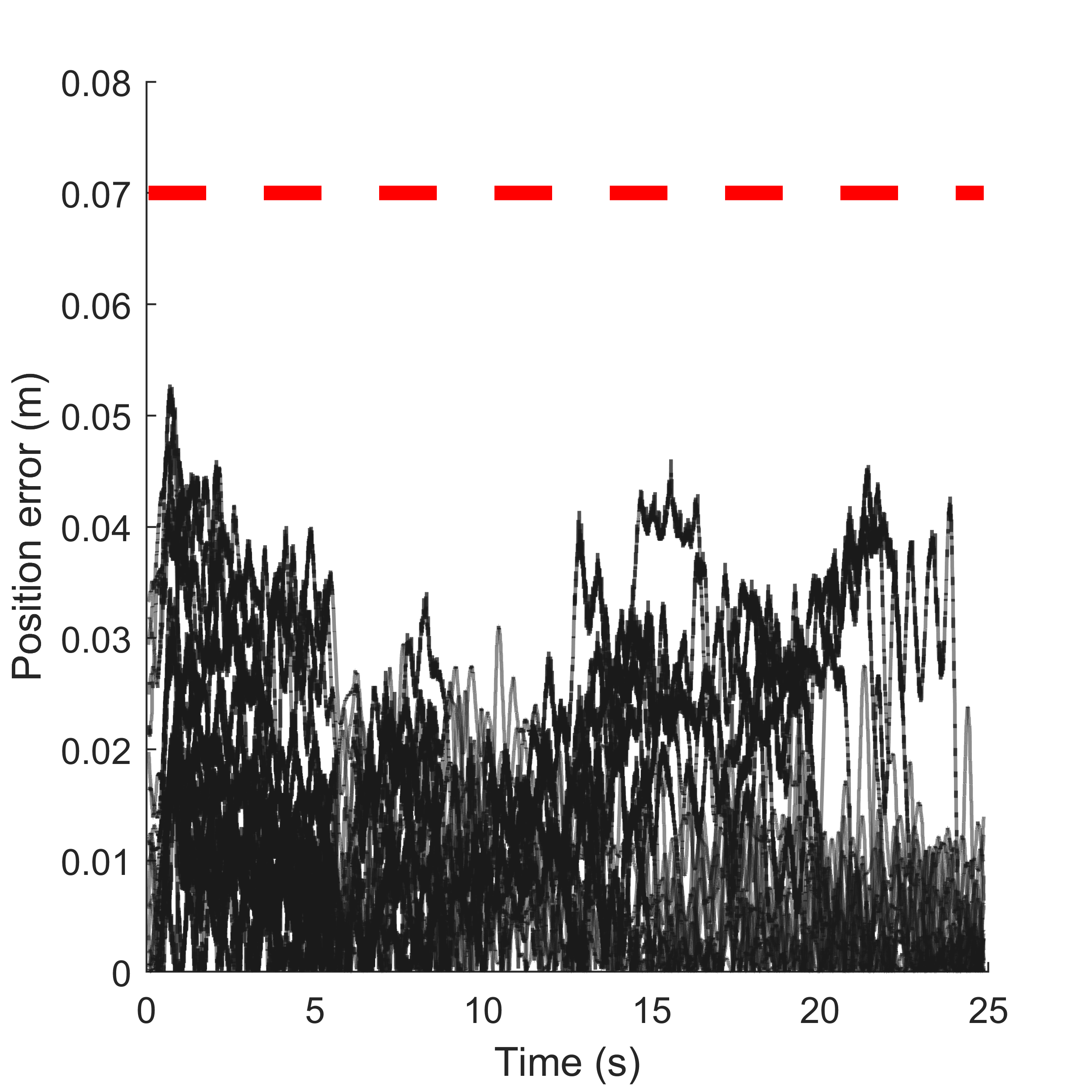}}  
  \caption{Position errors over time for all agents during the X20 experiment with (a) horizontal and (b) vertical components shown. Upper bounds used for trajectory planning are shown as dashed lines.} 
  \label{fig:position_error}  
\end{figure}

The results in Figures \ref{fig:experiment_photo}, \ref{fig:compare_sim_exp_O24_topview}, \ref{fig:compare_sim_exp_X20_clearance}, \ref{fig:position_error} demonstrate that the proposed method reliably generated trajectories that could be successfully tracked by a physical quadrotor team and executed in a reasonable time frame with guaranteed absence of collisions.

Videos demonstrating the experiment described in this paper as well as several others are available at \url{https://youtu.be/OapaAQAGWDE}.
The code which implements the algorithms described in this work and which supports both the virtual simulations and physical experiments can be found in \url{https://github.com/TSummersLab/cannon-tags}.

\section{Conclusion}
This work demonstrated tractable centralized methods for solving the goal assignment and inter-agent-collision-free trajectory planning problem for multiple robots. The assignment of agents to goals { achieved a low} total time-in-motion, and the resulting polynomial-in-time trajectories took full advantage of (possibly heterogeneous) speed capabilities. The results of numerical simulations revealed promising decreases in the total time with only mild increases in the computation time over existing approaches, allowing faster task completion in practical terms. The proposed algorithm also allowed us to eliminate restrictions present in other methods such as enforcement of synchronized start and end times and minimum separation of start and goal locations.

Future work is envisioned where the proposed framework would be used as a high-level centralized planner, combined with other decentralized techniques for dealing with lower-level local obstacles and disturbances.
The ability to use different altitudes i.e. all three spatial dimensions is crucial to the proper working of the proposed approach; operating spaces limited to a single 2D plane are not supported. Future work will investigate using curved (polynomial) paths to alleviate this issue while retaining tractability.

Future work also includes extension to agents with more complex dynamics and/or motion constraints, dealing with uncontrolled obstacles, combining time delays with altitudes, reassigning goals dynamically to further reduce would-be collisions, and a parallel implementation to decrease solve times. Investigation of the setting when there are more goals than agents and the setting of multiple stages is also warranted, both requiring dynamic goal assignment and replanning. 

Regarding the hardware implementation, refinements to the localization and state estimation furnished by the camera system as well as using more sophisticated controllers which account for downwash and ground effects \cite{Yeo2015,Shi2018} could further reduce the magnitude of the actual position errors and allow shrinkage of the collision volumes.

\clearpage
\appendix
\section{Algorithms}

\begin{algorithm} \label{algo:polynomial_minimization}
\DontPrintSemicolon
 \KwIn{Polynomial $p(t)$, interval $[t_0,t_f]$.}
    Differentiate $p(t)$ to obtain $\dot{p}(t)$ of degree $d-1$. This operation amounts to $d$ integer multiplications from the exponents of $t$.\;
     Find roots $t_i$ of $\dot{p}(t)$ by calculating eigenvalues of a companion matrix, as with the MATLAB \texttt{roots} function. \;
     Discard spurious solutions with imaginary part $\text{Im}(t_i) \neq 0$ or $t_i \notin [t_0,t_f]$. \;
     Append the endpoints $t_0$ and $t_f$ to the list $t_i$. \;
     Evaluate $p_i = p(t_i)$. \;
 \KwOut{Minimum $p^*=\min(p_i)$ and associated time $t^*=\text{argmin}(p_i)$.}
 \caption{Polynomial minimization}
\end{algorithm}

\begin{algorithm} \label{algo:polynomial_above}
\DontPrintSemicolon
 \KwIn{Polynomial $p(t)$, lower value $p_-$.}
     Find roots ${t_i}$ of $p(t)-p_-$. \;
     Discard spurious ${t_i}$ with imaginary part $\text{Im}({t_i}) \neq 0$. \;
     \eIf{$\{t_i\}=\emptyset$}{
        \eIf{$p(0) \geq p_-$}{
            ${\mathcal{T}_1}_- \gets [-\infty,\infty]$
        }{
            ${\mathcal{T}_1}_- \gets \emptyset$
        }
     }{
         Sort the roots ${t_i}$ in ascending order. \;
         \For{$i=1,2,\ldots,|\{{t_i}\}|+1$}{
             $t_{lwr} \gets {t_{i-1}}_-$ \;
             $t_{upr} \gets {t_i}$ \;
             $t_{eval} \gets ({t_{lwr} + t_{upr}})/2 $ \;
             \uIf{$i=1$}{
                $t_{eval} \gets {t_i} - 1$ \;
                $t_{lwr} \gets -\infty$ \;
             }
             \uElseIf{$i=|\{t_i\}|+1$}{
                $t_{eval} \gets {t_i} + 1$ \;
                $t_{upr} \gets \infty$ \;
             }
                Evaluate ${p_i}_- = p(t_{eval})$. \;
                \eIf{${p_i}_- \geq p_-$}{
                    ${\mathcal{T}_i}_- \gets [t_{lwr},t_{upr}]$ \;
                }{
                    ${\mathcal{T}_i}_- \gets \emptyset$ \;
                }
        }
     }
 \KwOut{Domain intervals ${\mathcal{T}_i}_- \neq \emptyset$.}
 \caption{Polynomial above value}
\end{algorithm}

\begin{algorithm} \label{algo:polynomial_interval}
\DontPrintSemicolon
 \KwIn{Polynomial $p(t)$, domain interval $\mathcal{T}_{\text{glob}} = [t_0,t_f]$, range interval $[p_-,p_+]$.}
    Find domain intervals ${\mathcal{T}_i}_-$ and ${\mathcal{T}_j}_+$ by calling Alg. \ref{algo:polynomial_above} with inputs $p(t)$ and $p_-$ or $p_+$ respectively. \;
    Initialize $k \gets 1$ \;
    \ForEach{Pair ${\mathcal{T}_i}_-$ and ${\mathcal{T}_j}_+$}{
        $\mathcal{T}_k \gets \{ {\mathcal{T}_i}_- \cap {\mathcal{T}_j}_+ \} \cap \mathcal{T}_{\text{glob}}$ \;
        $k \gets k+1$ \;
    }
 \KwOut{Time intervals $\mathcal{T}_k \neq \emptyset$.}
 \caption{Polynomial interval restriction}
\end{algorithm}

\begin{algorithm} \label{algo:segment_pair_collision}
\DontPrintSemicolon
 \KwIn{Heading unit vectors $\hat{h}_i$ and $\hat{h}_j$, polynomial trajectories $x_i(t) = p_i(t)\hat{h}_i$ and $x_j(t) = p_j(t)\hat{h}_j$ of degree $d$ over time intervals $\mathcal{T}_i = [t_{i,0},t_{i,f}]$ and $\mathcal{T}_j = [t_{j,0},t_{j,f}]$.}
    Calculate shared time interval $\mathcal{T}_{ij} = \mathcal{T}_i \cap \mathcal{T}_j$. \;
    \If{$\mathcal{T}_{ij}$ not empty}
    {
        \uIf{$\hat{h}_{i,3}=0$ and $\hat{h}_{j,3}=0$}{
            \If{\eqref{eq:vert_collide} true for $x_i(\tau)$, $x_j(\tau)$, $\tau \in \mathcal{T}_{ij}$}{
            Obtain $d^*$ from Alg. \ref{algo:separation_minimization} with inputs $\hat{h}_{i,12}$, $\hat{h}_{j,12}$, $x_i(t) = p_i(t)\hat{h}_{i,12}$, $x_j(t) = p_j(t)\hat{h}_{j,12}$, $\mathcal{T}_{ij}$.\;
                \If{\eqref{eq:horz_collide} true for $\|x_{j,12}-x_{i,12}\|=d^*$}{
                    \KwRet{True} \;
                }
            }
        }
        \uElseIf{$\hat{h}_{i,12}=0$ and $\hat{h}_{j,12}=0$}{
            \If{\eqref{eq:horz_collide} true for $x_i(\tau)$, $x_j(\tau)$, $\tau \in \mathcal{T}_{ij}$}{
            Obtain $d^*$ from Alg. \ref{algo:separation_minimization} with inputs $\hat{h}_{i,3}$, $\hat{h}_{j,3}$, $x_i(t) = p_i(t)\hat{h}_{i,3}$, $x_j(t) = p_j(t)\hat{h}_{j,3}$, $\mathcal{T}_{ij}$.\;
                \If{\eqref{eq:vert_collide} true for $|x_{j,3}-x_{i,3}|=d^*$}{
                    \KwRet{True} \;
                }
            }
        }
        \Else{
            Calculate the time interval(s) $\mathcal{T}_{ij,k}^\prime$ where \eqref{eq:vert_collide} is satisfied using Alg. \ref{algo:polynomial_interval}. \;
            \uIf{$\hat{h}_{i,3}=0$}{
                $\hat{h} = \hat{h}_{i}$
            }
            \Else{
                $\hat{h} = \hat{h}_{j}$
            }
            \ForEach{$\mathcal{T}_{ij,k}^\prime$}{
                Obtain boolean $b$ from Alg. \ref{algo:segment_pair_collision} with inputs \\ $\hat{h}$, $\hat{h}$, $x_i(t) = p_i(t)\hat{h}$, $x_j(t) = 0 \hat{h}$, $\mathcal{T}_{ij,k}^\prime$.\;
                \If{b}{
                \KwRet{True}
                }
            }
        }
    }
    \KwRet{False} \;
 \KwOut{Boolean of collision.}
 \caption{Segment pair collision check}
\end{algorithm}

\clearpage


\bibliographystyle{model1-num-names}

\bibliography{bibliography.bib}

\end{document}